\pdfoutput=1
\documentclass[journal]{IEEEtran}
\newcommand{\RR}{\mathbb{R}}

\newcommand{\roc}{\rm  ROC}

\newcommand{\cpauc}{\rm  CPAUC}
\newcommand{\auc}{\rm  AUC}

\def\E{{\cal E}}

\def\S{{\cal S}}

\def\X{{\cal X}}

\newtheorem{theorem}{Theorem}

\newtheorem{proposition}[theorem]{Proposition}

\newtheorem{rema}{Remark}
\newenvironment{remark}{\begin{rema} \rm}{\end{rema}}
\newtheorem{exam}{Example}

\usepackage{hyperref}

\usepackage{graphicx} 
\usepackage{subfigure} 
\usepackage{textcomp}
\usepackage{rotating}
\usepackage{euler}
\usepackage{amssymb}
\usepackage{amsmath}
\usepackage{amscd}
\usepackage{latexsym}
\usepackage{multirow}
\usepackage{lscape}
\usepackage{algorithm}
\usepackage{algorithmic}
\usepackage{rotating}
\usepackage{xcolor}
\ifCLASSINFOpdf
\else
\fi
\begin{document}
%
\title{Functional Bipartite Ranking: \\a Wavelet-Based Filtering Approach}
%
%
%

\author{St\'ephan~Cl\'emen\c{c}on
        and~Marine~Depecker~
\thanks{S. Cl\'emen\c{c}on is with the Department
of Image and Signal Processing, LTCI Telecom ParisTech/CNRS No. 5141, 75634 Paris, France e-mail: stephan.clemencon@telecom-paristech.fr.}
\thanks{M. Depecker is with the CEA, LIST, 91190 Gif-sur-Yvette, France.}
\thanks{Manuscript submitted December 2013}}

\maketitle

\begin{abstract}
It is the main goal of this article to address the \textit{bipartite ranking} issue from the perspective of functional data analysis (FDA). Given a training set of independent realizations of a (possibly sampled) second-order random function $X=(X(t))_{t\in [0,1]}$, valued in a path space $\X\subset \mathcal{L}_2([0,1])$, with a (locally) smooth autocorrelation structure and to which a binary label $Y\in\{-1,+1\}$ is randomly assigned, the objective is to learn a \textit{scoring function} $s:\X\rightarrow \mathbb{R}$ with optimal $\roc$ curve. Based on linear/nonlinear wavelet-based approximations, it is shown how to select compact finite dimensional representations of the input curves adaptively, in order to build accurate ranking rules, using recent advances in the ranking problem for multivariate data with binary feedback. Beyond theoretical considerations, the performance of the learning methods for functional bipartite ranking proposed in this paper are illustrated by numerical experiments.
\end{abstract}


%
\IEEEpeerreviewmaketitle

\section{Introduction}
\label{sec:intro}

\textit{Functional Classification}, \textit{i.e.} the binary classification problem when the input observation $X=(X(t))$ is of the form of a (possibly sampled) random curve/function and the output variable $Y\in\{-1,\; +1\}$ is a binary label, has been the subject of a good deal of attention in the machine-learning literature in the past few years, see \cite{RS02} or \cite{FV06}. In contrast, \textit{Bipartite Ranking}, termed \textit{Nonparametric Scoring} sometimes, has never been tackled in a functional framework, except from the restrictive angle of \textit{Functional Logistic Regression}, see \cite{RS97} or \cite{EAV04} for instance. This global learning task consists in ordering all possible input observations $X$ so that positive ones appear on top of the list with highest probability. This predictive problem, which can be cast in terms of $\roc$ curve optimization (see \cite{CV09ieee}), covers a wide variety of applications, ranging from anomaly detection in signal processing to automatic design of diagnosis tools in medicine through credit-scoring in mathematical finance or the conception of search engines in information retrieval.
\par Functional versions of many popular approaches for classification have been developed, relying in general on a preliminary finite dimensional representation/projection of the input data. When input observations consist of sampled curves on a fixed grid, \textit{Regularization Methods} are also used in order to handle problems related to the significant degree of autocorrelation of these high dimensional data. For instance, see \cite {HBT95} for the application of Linear Discriminant Analysis combined with regularization in the functional setting.
A widely-used alternative method in Functional Data Analysis (FDA), known as \textit{the filtering approach}, consists in projecting the input curves onto an adequate finite dimensional function subspace, the coefficients describing the latter being then used as input vectors, "feeding" next some classification learning algorithm for multivariate data. The basis functions are either selected through Principal Component Analysis (they correspond in this case to elements of the Karhunen-Loeve (KL) basis related to the input process $X$, supposedly of second order), see \cite{KS90} as well as \cite{HPP01} or \cite{RS02} for such a functional version of Quadratic Discriminant Analysis, or else are chosen among a dictionary of "time-frequency atoms" according to their capacity to represent efficiently the input process $X$, \textit{cf} \cite{CoifmanSaito}. One may refer to \cite{BBW06} for instance for a filtering stage based on the Fourier basis (before implementing $k-NN$ classification), to \cite{RV06} for use of spline interpolation (preliminary to SVM classification) and to \cite{BBR08} for a (nonlinear) wavelet-based filtering. 
\par Motivated by the increasing availability of functional data to ground scoring rules in various areas (\textit{e.g.} metabolomic/metabonomic measures collected by mass spectrometry or NMR techniques in biomedicine, return series in finance), it is the purpose of this paper to tackle bipartite ranking in a functional setting by means of wavelet filtering strategies combined with a recent tree-based ranking methodology termed {\sc TreeRank}, see \cite{CDV09}. Viewing bipartite ranking as a nested collection of binary classification problems with asymmetric costs, the major novelty arises from the fact that filtering is implemented \textit{locally} in an adaptive (nonlinear) and recursive manner, as permitted by the structure of the {\sc TreeRank} algorithm. We point out that this advantage is by no means confined to wavelet filtering and can be exploited with different techniques, including local regularization. As supported by strong empirical evidence, this approach may lead to more accurate scoring rules than those using filtering as a simple preprocessing step, especially when applied to stochastic processes with high variability at a wide range of resolution levels.
\par The paper is structured as follows. In Section \ref{background}, the bipartite ranking problem is rigorously formulated from a FDA perspective: basic concepts are briefly reviewed, notations are set out and the main assumptions required in the subsequent analysis are given. Section \ref{sec:filter} reviews a few theoretical results, showing that, even if wavelets are not the KL basis for the input process, linear/nonlinear wavelet approximation is very efficient for a wide variety of distributions. The extension of the {\sc TreeRank} algorithm to the functional setup we promote here is next described in Section \ref{sec:algo}, together with a discussion of practical issues related to its implementation. Results of illustrative numerical experiments are displayed in Section \ref{sec:exp}. Technical details are postponed to the Appendix.
\par Eventually, notice that a brief preliminary version of this work has been outlined in the conference paper \cite{CDSP11}. The present paper differs from it in its more developed theoretical analysis (learning rate bounds are established, paving the way for the design of automatic model selection procedures) and in its significantly augmented experimental part, displaying more illustrative experimental results, on real datasets in particular, and providing a more detailed description of practical implementation aspects.

\section{Background} \label{background} 
We briefly recall the crucial notions involved in the bipartite ranking task and introduce the required notations and assumptions.
  
  \subsection{Probabilistic Setup}
Here and throughout, $\mathcal{L}_2([0,1])$ denotes the space of square integrable (w.r.t. Lebesgue measure) functions $f:[0,1]\rightarrow \RR$,  $X=(X(t))_{t\in[0,1]}$ is a stochastic process, taking its values in a function space $\X\subset \mathcal{L}_2([0,1])$, and $Y$ is a random label in $\{-1,\; +1\}$ that describes a binary feedback ("relevant" \textit{vs.} "irrelevant", "healthy" \textit{vs.} "sick", "normal" \textit{vs.} "abnormal", \textit{etc.}). The stochastic process $X$ is regarded as a random observation for predicting the binary label $Y$. Let $p=\mathbb{P}\{ Y=+1 \}$ be the rate of positive instances. The joint distribution of the random vector $(X,Y)$ is characterized by the pair $(\mu, \eta)$, where $X$'s marginal distribution is denoted by $\mu$ and the \textit{posterior probability} by $\eta(x)=\mathbb{P}\{ Y=+1\mid X=x \}$, $x\in \X$. The indicator function of any event $\mathcal{E}$ is denoted by $\mathbb{I}\{\mathcal{E}\}$, the usual Hilbert norm on $\mathcal{L}_2([0,1])$ by $\vert\vert.\vert\vert$. Notice finally that the subsequent analysis can be easily extended to multivariate second-order random fields defined on bounded domains, see \cite{CDGO02}. For the sake of simplicity, focus is here on univariate stochastic processes.
  
\subsection{Functional Bipartite Ranking}\label{subsec:ranking}

Due to its ubiquity in the applications (signal processing, information retrieval, \textit{etc.}), the statistical and machine-learning communities have recently shown increasing interest in bipartite ranking these last few years, see \cite{CLV08} and the references therein for instance.

\subsubsection{$\roc$ analysis} In contrast to standard binary classification, where the goal is to guess, for a given $x\in \X$, the likeliest label $C^*(x)=2\cdot \mathbb{I}\{\eta(x)>1/2\}-1$, the ranking task consists in sorting all the instances $x\in \X$ by increasing order of the posterior probability $\eta(x)$. In practice, preorders on $\X$ are defined by means of measurable functions $s:\X\rightarrow \mathbb{R}$, generally termed \textit{scoring functions}: $\forall (x,x')\in \X^2$, $x\leq_s x' \Leftrightarrow s(x)\leq s(x')$. The ranking accuracy is generally measured in a functional manner, through the plot of its $\roc$ curve $t\in\mathbb{R}\mapsto (\mathbb{P}\{s(X)>t\mid Y=-1\}, \mathbb{P}\{s(X)>t\mid Y=+1\})$, see \cite{vTr68}. Connecting possible jumps by linear segments, the $\roc$ curve of $s$ can be viewed as the (continuous) graph of a function $\alpha\in (0,1)\mapsto \roc_s(\alpha)$. From the perspective of statistical hypothesis testing, the quantity $\roc_s(\alpha)$ can be seen as the power of the test of level $\alpha$ based on the statistic $s(X)$, in order to discriminate between the null assumption $\mathcal{H}_0:\; Y=-1$ and the alternative $\mathcal{H}_1:\; Y=+1$ . This notion naturally induces a partial order on the set $\S$ of all scoring functions: $s_1$ is said more accurate than $s_2$ iff $\roc_{s_2}(\alpha)\leq \roc_{s_1}(\alpha)$ for all $\alpha\in (0,1)$. Standard Neyman-Pearson arguments show that the dominating $\roc$ curve is that of scoring functions inducing the same ordering on $\X$ as $\eta(x)$, see Proposition 4 in \cite{CV09ieee} for instance. We denote it by $\roc^*(=\roc_{\eta})$. In order to remedy the partial nature of the preorder induced by the $\roc$ curve criterion on the set of scoring rules, a scalar performance measure, $\auc(s)=\int_{[0,1]}\roc_s(\alpha)d\alpha$, termed the \textit{area under the $\roc$ curve} (see \cite{HaMcNe82}), is widely considered in practice: $\forall s\in \S,\; \auc(s)\leq \auc^*\overset{def}{=}\auc(\eta)$. Its popularity mainly arises from its possible probabilistic interpretation as the \textit{rate of concording pairs}: for all $s\in \S$, we have
\begin{multline*}
\auc(s)=\mathbb{P}\left\{  s(X)<s(X')\mid (Y,Y')=(-1,+1) \right\}\\+\frac{1}{2}\mathbb{P}\left\{  s(X)=s(X')\mid (Y,Y')=(-1,+1)\right\},
\end{multline*}
where $(X',Y')$ denotes an independent copy of $(X,Y)$.

\subsubsection{Learning to rank data with binary feedback} Based on a "training" sample of i.i.d. labeled examples $\mathcal{D}_n=\{(X_1,Y_1),\;\ldots,\; (X_n,Y_n)\}$, the goal is to find a scoring function $\widehat{s}_n$ with a $\roc$ curve close to the optimal one $\roc^*$ or such that the "$\auc$-deficit" $\auc^*-\auc(\widehat{s}_n)$ is small, in order predict the order of new (unlabeled) observations as accurately as possible. For this purpose, various learning algorithms have recently been introduced in the literature, among which \cite{FISS03}, \cite{CV09CA} or \cite{PTABS07}. Focus is here on a flexible recursive method, termed {\sc TreeRank}, that produces interpretable and visualisable ordering rules of the form of \textit{oriented binary ranking trees}. Refer to \cite{CDV09} for further details on its implementation and to \cite{CV09ieee} for rigorous statistical foundations.

\subsubsection{Functional setting} Here, motivated by many applications (signal processing, spectrometry, \textit{etc.}), input observations $x\in \X$ are assumed to be realizations of the second-order stochastic process $X$.  Rather than replacing $X(t)$ by $X(t)-\mathbb{E}[X(t)]$, it is assumed in the sequel that the input process is centered for simplicity (notice nevertheless that the structure of the first order could easily be incorporated to the present study). Here and throughout, $X$'s covariance function is denoted by $C_X(s,t)=\mathbb{E}[X(t)X(s)]$. The properties of the random function $X$ involved in the subsequent analysis will be expressed in a probabilistic sense, in terms of (local) smoothness assumptions for $C_X$.

\section{Wavelet-Based Filtering}\label{sec:filter}
Dealing with infinite-dimensional data requires to reduce dimension. From the angle embraced in this paper, the \textit{filtering approach} is considered for this purpose. As shall be seen below, wavelet approximation may permit to control the loss of information, in the mean square sense, inherent in projecting the input curves onto a finite-dimensional subspace, provided it is defined by a sufficiently regular orthonormal wavelet basis.

\subsection{Filtering the Input Curves}
Let $(g_n)_{n\geq 1}$ be an orthonormal basis of $\mathcal{L}_2([0,1])$. A natural way of reducing dimension is to consider the projection of the (random) function $X$ onto a finite-dimensional subspace, spanned by a subcollection $\mathcal{G}_N$ of $N\geq 1$ basis functions, the first ones say $g_1,\;\ldots,\; g_N$,
\begin{equation*}
\mathcal{P}_{\mathcal{G}_N}X(t)=\sum_{n=1}^N\langle X, g_n \rangle g_n(t).
\end{equation*}
The basis that minimizes 
$$
\epsilon_N(X)\overset{def}{=}\mathbb{E}\left[\vert\vert X-\mathcal{P}_{\mathcal{G}_N}X\vert\vert^2\right]=\sum_{n>N}\mathbb{E}\left[\left(  \langle X, g_n \rangle\right)^2\right]
$$
 for every $N\geq 1$ is known as the Karhunen-Lo\`eve basis (KL). Since $\mathbb{E}[(  \langle X, g_n \rangle)^2]=\int\int C_X(t,s)g_n(t)g_n(s)dtds$, the KL basis can be shown to diagonalize the covariance operator $h(s)\mapsto \int_tC_X(t,s)h(t)dt$, see Theorem 9.3 in \cite{Mallat99}. 
 \par As the law of the process $X$ is unknown in general, the KL basis must be estimated based on a set of realizations of the process $X$. Given the extreme difficulty of this task, the use of function bases introduced in the field of \textit{Computational Harmonic Analysis} (CHA), that possess nice computational properties and approximately decorrelate a wide variety of stochastic processes both a the same time,  has become increasingly popular in the fields of signal/image processing and data compression. In particular, as shown in \cite{CDGO02} for instance, wavelets may provide compact representations of realizations of a wide variety of stochastic processes $X$. In order to describe next related linear and nonlinear approximation schemes, we introduce some additional notations. We denote by $(\psi,\;\phi)$ a pair of compactly supported wavelet/scaling function on $[0,1]$\footnote{For simplicity, we make no notational distinction between \textit{edge} and \textit{interior} wavelet (scaling function, respectively). Refer to \cite{CDV93} for further details on wavelet bases on the interval.} with $\mathcal{M}$ vanishing moments (\textit{i.e.} $\int t^k\psi(t)dt=0$ for $k\in\{0,\;\ldots,\; \mathcal{M}-1\}$).
For all $(j,k)\in \mathbb{N}\times \{0,\; \ldots,\; 2^j-1\}$, we set
\begin{eqnarray*}
\alpha_{j,k}&=&\int_{t=0}^1X(t)\phi_{j,k}(t)dt,\\
\beta_{j,k}&=&\int_{t=0}^1X(t)\psi_{j,k}(t)dt,
\end{eqnarray*}
with the notation $\gamma_{j,k}(t)=2^{j/2}\gamma(2^jx-k)$ for any function $\gamma:\mathbb{R}\rightarrow \mathbb{R}$.
We refer to Chapter VII of \cite{Mallat99} for computational aspects of wavelet transforms based on (uniform) sampling of square integrable signals. 

\subsection{Linear Wavelet Filtering} \label{subsec:lin}
Let $j\geq 0$. The projection of $X$ onto the subspace spanned by the $\phi_{j,k}$'s is given by:
\begin{eqnarray}\label{eq:approxlin}
\mathcal{P}_jX(t)&=&\sum_{k=0}^{2^j-1}\alpha_{j,k}\phi_{j,k}(t)\\
&=&\mathcal{P}_{j_0}X(t)+\sum_{m=j_0}^{j}\sum_{k=0}^{2^m-1}\beta_{m,k}\psi_{m,k}(t),\nonumber
\end{eqnarray}
This approximation uses $N_j=2^j$ terms. We denote by $\mathbf{X}_j$ the collection of random coefficients $\{\alpha_{j,k}:\; 0\leq k<2^j\}$ that can be used in order to represent the stochastic process $X$. For clarity, we recall the following result (see Proposition 2 in \cite{CDGO02}), which basically shows that such a linear approximation scheme is efficient in the case where $X$'s paths are uniformly smooth over $[0,1]$.

\begin{theorem} \label{thm:approx1} (\cite{CDGO02}) Let $0<r<(\mathcal{M}-1)/2$. Suppose that the autocorrelation function $C_X(t,s)$ is of class $\mathcal{C}^{2r}$ on the diagonal $\{t=s\}$\footnote{Recall that a function $F:[0,1]^2\rightarrow \mathbb{R}$ is said to be of class $\mathcal{C}^{2r}$ at $(t_0,s_0)$ iff there exists $c<\infty$ such that $\vert F(t,s)-P(t,s)\vert\leq c(\vert t-t_0\vert +\vert s-s_0\vert)$, where $P$ is a polynomial of degree $\lfloor r\rfloor$.}.
 Then, there exists $c<+\infty$ such that: for all $j\geq 1$,
\begin{equation*}
 \mathbb{E}\left[ \vert\vert X-\mathcal{P}_{j}X \vert\vert^2 \right]\leq c\cdot N_j^{-2r}.
\end{equation*}
\end{theorem}
This linear approximation bound simply follows from the fact that $\mathbb{E}[\alpha_{j,k}^2]$ coincides with the "diagonal coefficient" $\int\int C_X(t,s)\phi_{j,k}(t)\phi_{j,k}(s)dtds$ of the autocovariance function in the bivariate basis obtained by the method of tensorial product, combined with the supposed smoothness properties of $C_X$. Namely, using standard integration-by-part arguments, on may show that this implies: $\mathbb{E}[\alpha_{j,k}^2]\leq C \cdot 2^{-(2r+1)j}$, for all $(j,k)$. It is noteworthy that, under Theorem \ref{thm:approx1}'s assumptions, (linear) wavelet approximation attains the same rate as approximation in the KL basis, see the discussion in section 3 of \cite{CD97} for instance.

\subsection{Non Linear Wavelet Filtering}\label{subsec:nonlin}
Let $j_0\geq 0$ be fixed. Instead of considering the random curve $X$, we shall now consider a (nonlinear) wavelet-based approximation of the random function $X$ based on the $N$ wavelet coefficients with highest variance among those of resolution level larger than $j_0$.
In order to write the \textit{approximant} explicitly, we abusively set $\phi_{j_0,k}=\psi_{j_0-1,k}$ and $\beta_{j_0-1,k}=\alpha_{j_0,k}$ for $0\leq k<2^{j_0}$ and reindex the wavelet coefficients $\beta_{j,k}$, $j_0-1\leq j$ so that
$$
\mathbb{E}[\beta^2_{j(1),k(1)}]\geq \mathbb{E}[\beta^2_{j(2),k(2)}]\geq \cdots .$$
Equipped with this notation, we consider the "$N$-term approximant" obtained by keeping the terms corresponding to the coefficients with largest second-order moment and discarding the others:
\begin{equation}\label{eq:approx}
\widetilde{\mathcal{P}}_{N}X(t)=\sum_{(j,k)\in \mathcal{I}_N} \beta_{j,k}\psi_{j,k}(t),
\end{equation}
where $\mathcal{I}_N=\{(j(l),k(l)):\; 1\leq l\leq N\}$. The terms at the lowest scale should be seen as the "gross structure" and the others as "details" refining the accuracy of the approximation. We denote by $\widetilde{\mathbf{X}}_{N}$ the related collection of coefficients. For all subset of indices $\mathcal{J}\subset \{(j,k): \; j\geq j_0-1,\; 0\leq k <2^j\}$ of cardinality $\# \mathcal{J}<+\infty$, we denote by $\pi_{\mathcal{J}}:\X\rightarrow \mathbb{R}^{\#\mathcal{J}}$ the filter that assigns to any curve $x\in \X$ the collection of coefficients $(\beta_{j,k})_{(j,k)\in \mathcal{J}}$. Equipped with this notation, we have: $\widetilde{\mathbf{X}}_{N}=\pi_{\mathcal{I}_N}(X)$.

\par As the quantities $\mathbb{E}[\beta^2_{j,k}]$ are generally unknown, in practice we compute the statistical versions based on the observed paths $X_1,\;\ldots,$ $X_n$, 
$$
\frac{1}{n}\sum_{i=1}^n\left(\int X_i(t)\psi_{j,k}(t)dt\right)^2,$$
up to a maximum level of resolution $j_{\max}$ and retain the $N$ wavelet coefficients with highest empirical variance. This aspect of the approximation method shall be neglected in the subsequent theoretical analysis for simplicity's sake.
\par Notice that the approximant (\ref{eq:approx}) is different from the one that is considered in \cite{CDGO02} for instance (see Eq. $(8)$ therein), the index set involved in the definition of the latter depending on the path $X$. As revealed by the result below, the $N$-term approximation (\ref{eq:approx}) provides an efficient way of representing $X$ in the $\mathcal{L}_2$ sense.

\begin{theorem} \label{thm:approx2} Let $0<r<\mathcal{M}$. Suppose that the (random) wavelet coefficients obey the following constraint:
\begin{equation}\label{eq:constr}
\sum_{j\geq j_0-1}\sum_k\left(\mathbb{E}[\beta_{j,k}^2]\right)^{p/2}<+\infty,
\end{equation}
for $p=2/(1+2r)$. There exists $c<+\infty$ such that
\begin{equation*}
\forall N\geq 1,\;\; \mathbb{E}\left[ \vert\vert X-\widetilde{\mathcal{P}}_{N}X \vert\vert^2 \right]\leq c\cdot N^{-2r}.
\end{equation*}
\end{theorem}
Condition (\ref{eq:constr}) can be expressed in terms of diagonal wavelet coefficients of $X$'s covariance $C_X(u,v)=\mathbb{E}[X(u)X(v)]$, since $\mathbb{E}[\beta_{j,k}^2]=\int\int C_X(u,v)\psi_{j,k}(u)\psi_{j,k}(v)dudv$, and related to the smoothness of $C_X$ near the diagonal $\{u=v\}$, see \cite{CDGO02}. Typically, the "Besov constraint" $\mathcal{B}^{2r}_{2,2}$ (\ref{eq:constr}) describes objects whose smoothness is not "stationary". In particular, stochastic processes that fulfill assumption (\ref{eq:constr}) are those whose autocovariance function is \textit{piecewise smooth} near the diagonal line, \textit{i.e.} jumps at a few occasional points and is smooth in between. For such processes, it can be shown that this approximation scheme outperforms any type of linear approximation (\textit{cf} subsection \ref{subsec:lin}), including that based on the KL basis. One may refer to \cite{CDGO02} and \cite{CD97} for further details on the efficiency of nonlinear wavelet approximation.

\subsection{Bias \textit{vs.} Dimension Reduction}\label{subsec:bias}
The next result shows how optimal $\auc$ is affected by the dimension reduction method described above. A small distortion (measured by the $\mathcal{L}_2$ norm), as that induced by (linear/nonlinear) wavelet approximants in some situations, cannot lead to a large decrease of the optimal $\auc$.

\begin{proposition}\label{prop:bias}{\sc (Distortion rates)} Assume that the regression function $\eta:\X\rightarrow [0,1]$ is Lipschitz.
\begin{itemize}
\item[(i)] Suppose that Theorem \ref{thm:approx1}'s assumptions are fulfilled. There exists a constant $c<\infty$ such that:
\begin{equation}
\forall j\geq 0,\;\; \auc^*-\auc^*_j\leq c\cdot N_j^{-r},
\end{equation}
where $N_j=2^{j}$ and $\auc^*_j$ denotes the maximum $\auc$ over the set of scoring functions defined on the space $\X_j$ of dimension $N_{j}$ in which the r.v. $\mathcal{P}_{j}X$ (or equivalently, $\mathbf{X}_{j}$) is valued.
\medskip

\item[(ii)] Suppose that Theorem \ref{thm:approx2}'s assumptions are fulfilled. There exists a constant $c<\infty$ such that:
\begin{equation}
\forall N\geq 1,\;\; \auc^*-\widetilde{\auc}^*_N\leq c\cdot N^{-r},
\end{equation}
where $\widetilde{\auc}^*_N$ denotes the maximum $\auc$ over the set of scoring functions defined on the space $\widetilde{\X}_J$ of dimension $N$ in which the r.v. $\widetilde{\mathcal{P}}_{N}X$ (or equivalently, $\widetilde{\mathbf{X}}_{N}$) takes its values.
\end{itemize}
\end{proposition}

 Before showing how the wavelet filtering methods impact on the learning rates of empirical risk minimization (ERM) methods (or, more precisely, of $\auc$ maximization techniques in the present case), a few remarks are in order.
\begin{remark} ({\sc On the Lipschitz assumption.}) We point out that the Lipschitz condition is fulfilled by a variety of generative models, including for instance the \textit{nonparametric logit linear model} given by:
$$
\log \left( \frac{\eta(x)}{1-\eta(x)}\right)=a+\int_{t=0}^1b(t)x(t)dt,
$$
where $(a,b)\in \mathbb{R}\times \mathcal{L}_2([0,1])$.
We also underline that, following in the footsteps of the analysis carried out in Section 32.2 of \cite{DGL96}, one may show for instance that $\auc_j^*\rightarrow \auc^*$ as $j\rightarrow \infty$ under Theorem \ref{thm:approx1}'s assumptions without requiring the Lipschitz property for $\eta(x)$.
\end{remark}

\begin{remark}\label{rk:shrink}({\sc Alternative nonlinear filtering schemes.}) Notice that, filtering the training sample paths $X_1,\;\ldots,\; X_n$ by keeping the $N$ coefficients with largest second order moment is by no means the sole nonlinear filtering strategy possible. \textit{Wavelet shrinkage}, a celebrated (level-dependent) thresholding scheme, can also be considered for this purpose. For instance, in the case where Theorem \ref{thm:approx2}'s hypotheses are satisfied, applying a hard thresholding at level $\sqrt{j/N^{r+1}}$ to coefficients $\mathbb{E}[\beta^2_{j,k}]$ (\textit{i.e.} replacing $\beta_{j,k}$ by $\beta_{j,k}\cdot \mathbb{I}\{\mathbb{E}[\beta^2_{j,k}]\geq \sqrt{j/N^{r+1}}\}$) for $j\leq c \log N /\log 2$ leads to an approximation with $O(N)$ coefficients that yields a distortion rate of the same order as \eqref{eq:approx}, see \cite{DJKP95} for further details.
\end{remark}

\subsection{$\auc$ Consistency of ERM Methods - Model Selection}\label{subsec:select}
It should be noticed that Proposition \ref{prop:bias} may permit to compute explicitly learning rates in the ranking context, that take into account the bias inherent in the dimension reduction step. For instance, combining it with Proposition 6 in \cite{CDV09} and Corollary 3 in \cite{CLV08}, the expected $\auc$ deficit of a scoring rule with minimum empirical $\auc$, such as those built by the {\sc TreeRank} method implemented with the standard {\sc CART} algorithm as {\sc LeafRank} procedure (more details are given in Subsection \ref{subsec:TreeRank}) and fed with the $N$-dimensional representation (\ref{eq:approx}) for the training data, is of order $\max\{ \sqrt{\log N/n},\; N^{-r}\}$ under appropriate assumptions, see \cite{CDV09}. This shows that choosing the dimension in a way that $N\sim n^{1/(2r)}$ leads to an optimal trade-off between bias and stochastic error in this case, and to a rate of order $\sqrt{\log n/n}$.

\par However, the nature and degree of smoothness of the autocorrelation function $C_X$ (near the diagonal) is generally unknown to the statistician. From a practical perspective, one has to rely only on the data to select the \textit{complexity parameter} $N$, assumed to belong to a range $\{N_{\min},\;\ldots,\; N_{\max}\}\subset \mathbb{N}^*$, specified in advance. Following a popular approach in automatic model selection \cite{Dev88}, we now show how to penalize the empirical counterpart of the $\auc$ criterion by adding a complexity term $pen_n(N)$ (increasing with the dimension $N$) for this purpose. In order to describe the selection strategy precisely, we introduce $\mathcal{S}_N$, a {\sc VC} major class of (scoring) functions $S:\mathbb{R}^N\rightarrow \mathbb{R}$ of finite {\sc VC} dimension $V_N$ (see \cite{Dud99}), among which a scoring rule $\widehat{S}_{N,n}$ is selected, in order to maximize the statistical version of the $\auc$ criterion over $\S_N$:
$$
\widehat{AUC}_n(S\circ\pi_{\mathcal{I}_N}),
$$
where, for all $s\in \S$, $\widehat{AUC}_n(s)$ is given by
\begin{multline*}
\frac{1}{n_+ n_-}\sum_{1\leq i\neq j\leq n} \mathbb{I}\left\{ s(X_i))< s(X_j)),\; (Y_i,Y_j)=(-1,+1) \right\}\\
+\frac{1}{2}\frac{1}{n_+n_-}\sum_{1\leq i\neq j\leq n}\mathbb{I}\left\{ s(X_i))< s(X_j)),\; (Y_i,Y_j)=(-1,+1) \right\},
\end{multline*}
denoting by $n_+=n-n_-=\sum_{i=1}^n\mathbb{I}\{Y_i=+1\}$ the number of positive curves among the training dataset and by $\pi_{\mathcal{I}_N}$ the adaptive non linear wavelet filter described in Subsection \ref{subsec:nonlin}.
For all $x\in \X$, we set $\widehat{s}_{N}(x)=(\widehat{S}_{N,n}\circ\pi_{\mathcal{I}_N})(x)$. Practical methods for (approximate) $\auc$ optimization are not described in this Subsection (see the references listed in Section \ref{sec:algo}), focus is here on the statistical properties of the complexity penalized empirical $\auc$ maximizers. In order to avoid overfitting, we choose $N_{opt}$ in $\{N_{\min},\;\ldots,\; N_{\max}\}$ so as to maximize the \textit{complexity penalized empirical $\auc$ criterion}:
$$
\widehat{\cpauc}_{n}(\widehat{s}_{N_{opt}})=\max_{N_{\min}\leq N\leq N_{\max}}\widehat{\cpauc}_{n}(\widehat{s}_N),
$$
where $\widehat{\cpauc}_n(s)=\widehat{\auc}_n(s)-pen_n(N)$ for all $s\in \S$.

\par The following result reveals that, for a suitable choice of $pen_n(N)$, the selection procedure automatically finds the best balance between the approximation error (bias) inherent in the choice of the dimension $N$ and a {\sc VC} dimension-based (distribution-free) upper bound for the estimation error. 

\begin{proposition}\label{prop:select}{\sc (Automatic dimension reduction)} For all $N\in \{N_{\min},\;\ldots,\; N_{\max}\}$, take as dimension penalty:
$$
pen_n(N)=4\sqrt{\frac{V_N\log(n+1)+\log 2}{n}}.
$$
 Then, we have: $\forall n\geq 1$,
\begin{multline*}
\auc^*-\mathbb{E}[\auc(\widehat{s}_{N_{opt}})]\leq \sqrt{\frac{1}{2n}}+\\ \min_{N_{\min}\leq N\leq N_{\max}}\left\{\auc^*-\widetilde{\auc}^*_N +
pen_n(N)+\sqrt{\frac{\log N}{n}}\right\}.
\end{multline*}
\end{proposition}
\begin{remark}({\sc Examples.})
For the collection of (tree-structured) piecewise constant collection of scoring functions on $\mathbb{R}^N$ considered in \cite{CDV09} (see Subsection 4.2 therein), the {\sc VC} dimension $V_N$ is proportional to the feature space dimension $N$. When condition \eqref{eq:constr} is fulfilled for some unknown parameter $r$, the structural $\auc$ maximization technique described above yields the same rate as that which would have been obtained with the help of an oracle revealing us the value of $r$ (therefore the optimal choice $N\sim n^{1/(2r)}\in [N_{\min},\;N_{\max}]$).
\end{remark}

\begin{remark}({\sc Alternative model selection methods.}) The dimension selection strategy above, in the spirit of \textit{structural risk minimization} \cite{Vapnik98}, classically relies on an approximate upper bound for the stochastic term $\sup_{S\in \S_N}\vert \widehat{\auc}(S\circ \pi_{\mathcal{I}_N})-\auc(S\circ \pi_{\mathcal{I}_N}) \vert$ based on $\S_N$'s {\sc VC} dimension. We point out that similar results can be obtained using different upper bounds, relying on alternative notions of complexity for $\S_N$ (possibly depending on the data, see \cite{CLV08}), in terms of Rademacher averages for instance, or on an extra validation sample (as in \cite{BBR08} in the context of functional classification). Refer to \cite{Lug02} for further details.
\end{remark}
\section{{\sc TreeRank}: a Functional Version} \label{sec:algo}

In \cite{CV09ieee} and \cite{CDV09}, a novel recursive partitioning method, termed {\sc TreeRank} and producing tree-structured piecewise constant scoring functions has been introduced. We start with a brief description of the principles of this $\roc$ optimization technique and next show how one may take advantage of the latter in the infinite dimension framework, so as to design a very flexible extension of {\sc TreeRank} to the functional setting, where non linear wavelet filters are implemented \textit{locally}.

\subsection{The Standard {\sc TreeRank} Algorithm}\label{subsec:TreeRank}

\begin{figure}[t]
\centering
\includegraphics[width=8cm]{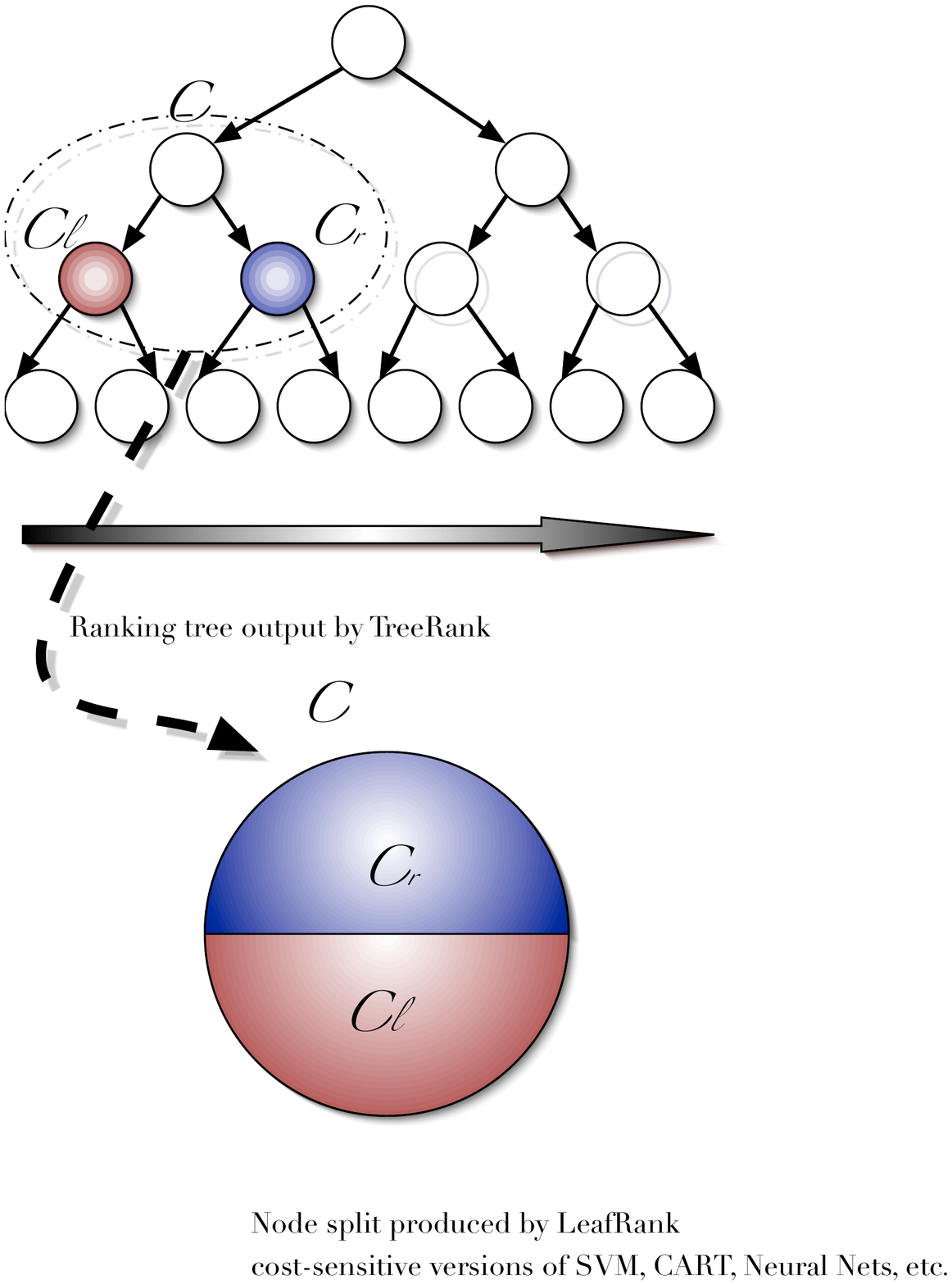}\vspace{-2cm}
\caption{Visualization of the ranking rule produced by {\sc TreeRank} by means of an oriented tree schematic.\label{fig:TreeRank}}
\end{figure}

 The {\sc TreeRank} algorithm produces an oriented partition of the feature space $\X$, whose structure can be represented by a left-to-right oriented binary tree, termed \textit{ranking tree}, see Fig. \ref{fig:TreeRank}. The root node corresponds to the whole space $\X$, each internal node represents a specific set $\mathcal{C}\subset\X$, while its left and right siblings correspond to subsets, $\mathcal{C}_l$ and $\mathcal{C}_r$ respectively, that form a partition of $\mathcal{C}$. The left-to-right orientation naturally defines an ordering of the terminal cells, therefore a preorder on the feature space, input observations lying in the same terminal cell being tied. 
 
\par The ranking tree is learnt from training data $\mathcal{D}_n=\{(X_1,Y_1),\;\ldots,\; (X_n,Y_n)\}$ in two stages. The \textit{growing stage}
 consists in calling recursively a \textit{binary classification algorithm with asymmetric cost}, termed {\sc LeafRank} in a generic manner. Given a pair of random vectors $(X',Y')$, valued in a space $\mathcal{C}\times\{-1,+1\}$,
 a cost $\omega\in (0,1)$ and $m\geq 1$ independent copies $(X'_1,Y'_1),\;\ldots,\; (X'_m,Y'_m)$ of $(X',Y')$, such an algorithm aims at splitting the cell $\mathcal{C}$ into subcells $\mathcal{C}_l$ and $\mathcal{C}_r=\mathcal{C}\setminus \mathcal{C}_l$ so as to minimize the \textit{empirical weighted misclassification risk}
\begin{multline*}
\widehat{L}_{\mathcal{C},\omega}(\mathcal{C}_l)=\frac{2}{m}(1-\omega)\sum_{i=1}^m\mathbb{I}\{X'_i\in \mathcal{C}_r,\; Y_i'=+1\}\\
+ \frac{2}{m}\omega\sum_{i=1}^m\mathbb{I}\{X'_i\in \mathcal{C}_l,\; Y_i'=-1\}
\end{multline*}
or an approximate (possibly penalized/convexified) version. When considering multivariate data, various techniques can be used for this purpose, such as \textit{Support Vector Machines} or \textit{Classification Trees}, see \cite{FriedHasTib09} for a recent account of (cost-sensitive) classification methods. Having fixed a maximum depth $D\geq 0$ for the ranking tree or a minimum number of instances in a cell, below which one stops splitting, the growing stage of the {\sc TreeRank} method consists in applying the weighted classification algorithm chosen in a recursive manner, using the rate of positive instances within the cell $\mathcal{C}$ to split, namely
$$
\omega_{\mathcal{C}}=\frac{1}{m_{\mathcal{C}}}\sum_{i=1}^n\mathbb{I}\left\{X_i\in \mathcal{C},\;  Y_i=+1  \right\}
$$
with $m_{\mathcal{C}}=\sum_{i=1}^n\mathbb{I}\{X_i\in \mathcal{C}\}$, as cost coefficient and the $m_{\mathcal{C}}$ data lying in $\mathcal{C}$ as training sample. As for most other recursive partitioning methods, the growing stage may be followed by a \textit{pruning stage}, where children of a same parent node can be merged recursively in order to maximize an estimate of the $\auc$ criterion, based on cross-validation for instance, see \cite{CDV09} for further details. As shown in \cite{CV09ieee}, this algorithm can be viewed as a statistical version of an adaptive piecewise-linear interpolation scheme of the optimal $\roc$ curve (assumed of class $\mathcal{C}^2$ on $[0,1]$), leading to an approximant that is a broken line connecting knots $\{(\alpha_i,\roc^*(\alpha_i)):\; i=0,\;\ldots,\; K+1\}$ with $\alpha_0=0<\alpha_1<\ldots<\alpha_{K+1}=1$ and coincides with the $\roc$ curve of a piecewise constant scoring function
\begin{eqnarray*}
s^*_K(x)=\sum_{k=1}^{K+1}(K-k+1)\cdot \mathbb{I}\{x\in \mathcal{E}_{\alpha_{i-1},\alpha_{i}} \}
\end{eqnarray*}
built from a collection of (adaptively chosen) bilevel subsets $\mathcal{E}_{\alpha_{i-1},\alpha_{i}}=\{x\in \X:\; Q^*(\alpha_{i})\leq \eta(x)<Q^*(\alpha_{i-1})\}$, with $1\leq i \leq K+1$ and denoting by $Q^*(\alpha)$ the quantile of order $1-\alpha$ of the conditional distribution of $\eta(X)$ given $Y=-1$. It leads to consistent ranking rules $\widehat{s}_n$, in the sense that $\roc_{\widehat{s}_n}\rightarrow \roc^*$ as $n\rightarrow \infty$ in a pointwise manner provided that the weighted classification stages are performed accurately enough. As explained at length in \cite{CDV09}, the key to the efficiency of the {\sc TreeRank} algorithm is the capacity of the {\sc LeafRank} method to capture well the geometry of the $\mathcal{E}_{\alpha_{i-1},\alpha_{i}}$'s. As shall be seen below, the fact that {\sc TreeRank} can be interpreted as a recursive statistical recovery of bi-level sets of the regression function is its main appealing feature in regards to the extension to the functional setup.

\subsection{Local Adaptive Filtering - The Algorithm.}\label{subsec:algo} Following the filtering paradygm in FDA, a natural manner of extending this learning method to situations where training observations are curves with binary labels is to consider a specific collection of wavelet coefficients as input data (those with highest empirical variance namely). Here, exploiting the recursive partitioning principle on which {\sc TreeRank} is based, we propose to implement the nonlinear wavelet filtering locally, \textit{i.e.} to select the indexes of the coefficients retained based on the labeled curves lying in the node to be split. Indeed, while certain bilevel sets $\mathcal{E}_{\alpha,\alpha'}$ can be well approximated by regions of the function space $\X$ defined in terms of the values taken by a specific finite subcollection $\mathbf{X}$ of wavelet coefficients, certain others may not (see the toy example described in Section \ref{sec:exp}). The learning algorithm below permits, to a certain extent, to avoid dramatical errors due to this phenomenon (worsened by the hierarchical structure of the algorithm and the global nature of the ranking task, see \cite{CDV09} for a thorough discussion), in contrast to approaches where filtering is implemented beforehand, as a simple preprocessing step.\\
Suppose that a classification algorithm (with assymmetric cost) $\mathcal{A}$ is given. Fix $N\geq 1$, $D\geq 1$ and $j_{\max}\geq 0$. The FDA version of {\sc TreeRank} we propose here is implemented iteratively as shown below.\\

\bigskip

\begin{center}
{\sc The Functional TreeRank Algorithm}
\end{center}
\medskip

\begin{itemize}
\item ({\sc Initialization.}) Take $\X$ as root of the oriented binary tree: $C_{0,0}=\X$. Compute the scaling and wavelet coefficients $\beta_{j,l}^{(i)}$, $j_0-1\leq j\leq j_{\max}$, of each curve $X_{i}$.
\medskip

\item ({\sc Iterations.}) For $d=0,\ldots, D-1$, $k=0,\ldots, 2^d-1$,
\medskip

\begin{enumerate}
\item ({\sc Local cost.}) Compute the rate of positive curves within the cell $C_{d,k}$:
$$
\omega_{C_{d,k}}=\frac{1}{n}\sum_{i=1}^n\mathbb{I}\{X_i\in C_{d,k},\; Y_i=+1\}.
$$
\item ({\sc Local filtering.}) Keep the $N$ wavelet coefficients $\mathbf{X}_{d,k}$  with highest (local) second-order moment
$$
\frac{1}{\# \mathcal{C}_{d,k}}\sum_{i\in \mathcal{C}_{d,k} }\left(\beta_{j,k}^{(i)}\right)^2,
$$
denoting by $\# \mathcal{C}_{d,k}$ the cardinality of $\{i:\,\, X_i\in \mathcal{C}_{d,k}\}$.
\medskip

\item ({\sc Cost-sensitive classification.}) Based on the training data $\{(\mathbf{X}^{(i)}_{d,k},Y_i):\; 1\leq i \leq n, \;\; X_i\in \mathcal{C}_{d,k} \}$ and using algorithm $\mathcal{A}$, solve the weighted binary classification problem related to the finite-dimensional input space $\mathcal{X}'=\{ \mathbf{X}_{d,k}:\, X\in C_{d,k}\}$, the asymmetric cost $\omega=\omega_{C_{d,k}}$.
\medskip

\item ({\sc Cell split.}) Set $C_{d+1,2k}=\X'_+$ and $C_{d+1,2k+1}=C_{d,k}\setminus C_{d+1,2k}$.
\end{enumerate}
\medskip

\item({\sc Output.}) Build the piecewise-constant function
$$
s_D(x)=\sum_{k=0}^{2^D-1}(2^D-k+1)\cdot \mathbb{I}\{X\in C_{D,k}\}.
$$
\end{itemize}

\bigskip

The scoring rule output by the algorithm can be stored in a heap data structure, representing a complete rooted oriented binary tree, where a collection $\mathcal{I}_{d,k}$ of $N$ wavelet indices $(j,l)$ and a classifier $g_{d,k}:\X_{d,k}\rightarrow \{-1,+1\}$ defined on the set 
$\X_{d,k}\overset{def}{=}\{\pi_{\mathcal{I}_{d,k}}(x):\; \; x\in \X\}$
of wavelet coefficients with indexes in $\mathcal{I}_{d,k}$  are assigned to each inner node $(d,k)$, $0\leq d<D$ and $0\leq k <2^d$: the decision function $g_{d,k}(\pi_{\mathcal{I}_{d,k}}(x))$ takes the value $+1$ when an element $x$ of the cell related to the node $(d,k)$ belongs to its \textit{left child} (corresponding to node $(d+1, 2k)$), and the value $-1$ otherwise. 
\par Reflecting the fact that the smoothness of $X$'s trajectory may depend on the order of magnitude of the posterior probability $\eta(X)$, the collection of wavelet basis functions used to represent the signal thus depends on the node in which it lies.
\par We finally point out that consistency of the {\sc Functional TreeRank} algorithm and learning rates can be straightforwardly proved by combining the results obtained for the finite dimensional situation in \cite{CV09ieee} and \cite{CDV09} with those established in subsections \ref{subsec:bias} and \ref{subsec:select} under adequate assumptions.

\subsection{Practical Issues - Variants and Model Selection}
The method sketched above can be refined in several ways. As for the standard version (see Subsection \ref{subsec:TreeRank}), the growing stage may be followed by a pruning stage, where children of a same parent node can be merged recursively in order to maximize a cross-validation based estimate of the $\auc$ criterion. In addition, the number $N$ of wavelet coefficients retained at each split step can depend on the current node $(d,k)$ and picked in order to minimize the local weighted misclassification risk estimated by means of a standard data splitting device, following in the footsteps of \cite{BBR08}.In this case as well, the number of local features involved in the split rule can be determined by means of a model selection procedure, relying either on an oracle bound such as that stated in Proposition \ref{prop:select} in the spirit of \textit{structural risk minimization} or else on resampling techniques (\textit{e.g.} cross-validation).

The choice of the wavelet basis can also impact on the performance of the method (see Section \ref{sec:exp} below). In practice, it should be based on an estimate of the $\auc$, computed by means of an extra validation sample or through resampling techniques.
Finally, {\bf b}ootstrap {\bf agg}regat{\bf ing} techniques (\textit{bagging}, in abbreviated form), based on (pseudo-) metrics on sets of rankings, can be used in order to increase the stability of the scoring rules output by {\sc Functional TreeRank}, exactly like in the finite dimensional framework. Refer to \cite{CDV10b}, \cite{CDV12} and \cite{CDV12bis} for further details.

\section{Numerical Experiments}\label{sec:exp}

Here we illustrate, through numerical experiments, the concepts previously introduced. In particular, based on synthetic and real datasets, we evaluate the performances of the scoring rule output by the {\sc Functional TreeRank} algorithm described in Subsection \ref{subsec:algo}. We also compare these, on the synthetic toy example, to that of its natural competitor, built by implementing a preliminary wavelet filtering. Notice that the results stated in subsections \ref{subsec:bias} and \ref{subsec:select} set statistical grounds for the validity of this two-stage approach.

\subsection{Toy Examples}

We consider numerical examples, where the conditional distributions $G(dx)$ and $H(dx)$ are both mixtures of $K\geq1$ probability laws $F_1,\;\ldots,\; F_K$, with disjoint supports $\X_1,\;\ldots,\; \X_K$. In this case, the class distributions can be expressed as follows:
\begin{eqnarray*}\label{ClassDistrib}
G(dx) &=& \Sigma_{k=1}^K \omega^+_k \cdot F_k(dx),\\
H(dx) &= &\Sigma_{k=1}^K \omega^-_k \cdot F_k(dx),
\end{eqnarray*}
where the $\omega^+_k$'s and the $\omega^-_k$'s are two collections of nonnegative coefficients summing to one and such that:
\begin{equation}
\frac{\omega^+_1}{\omega^-_1}\geq \frac{\omega^+_2}{\omega^-_2}\geq \cdots \geq \frac{\omega^+_K}{\omega^-_K}~.
\end{equation}
In this situation, it can be easily seen that the sets $\X_1,\;\ldots,\; \X_K$ define nested sublevel sets of the (piecewise constant) regression function. An optimal scoring function (in the $\roc$ curve sense) is thus given by
\begin{equation}\label{OptSco}
s^*(X) = \sum_{k=1}^K (K-k+1)\cdot \mathbb{I}\{X\in \X_k\},
\end{equation}
whose $\roc$ curve is the piecewise linear curve connecting the knots 
$\{(0;0),\cdots,(\sum_{\kappa=1}^k \omega^-_\kappa;\sum_{\kappa=1}^k \omega^+_\kappa),\cdots, (1;1)\}$.

\begin{figure}[h!]
\begin{center}
\includegraphics[width =8cm]{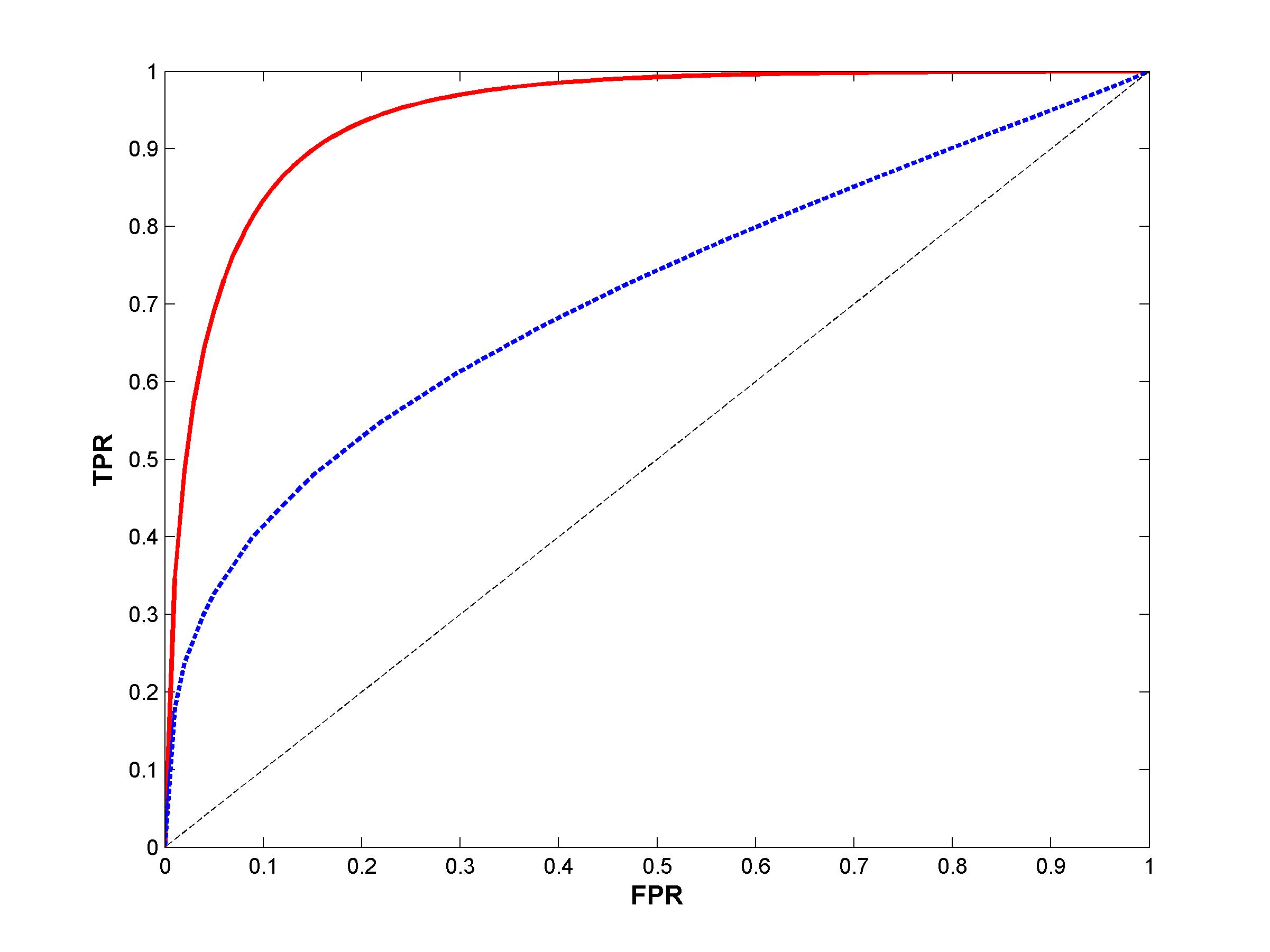}
\caption{\small{Optimal $\roc$ curves. {\itshape{Case a}} in red ($\auc^* = 0.94$) and {\itshape{Case b}} in blue dots ($\auc^* = 0.71$).}\label{fig:rocsopt}}
\end{center}
\end{figure}

Two examples are considered below, {\itshape{Case a}} and {\itshape{Case b}}, with $K=50$ mixture components. In each case, data have been generated in two stages as follows. The first step of the simulation consists in selecting the collections of coefficients $\Omega^+ = \{\omega_1^+,\;\ldots,\;  \omega_K^+\}$ and $\Omega^- =  \{\omega_1^-,\;\ldots,\;  \omega_K^-\}$. The choices made here lead to the curves displayed in Fig. \ref{fig:rocsopt}: clearly, negative and positive mixing coefficients, and thus class distributions, are much more similar in {\itshape{Case b}} than in {\itshape{Case a}}, making the discrimination problem harder: the optimal $\auc$ value is equal to $0.94$ in {\itshape{Case a}}, and to $0.71$ in {\itshape{Case b}}. Both collections give respectively the proportions of positive and negative instances within each level set and thus characterize the toughness of the ranking problem. In both examples, the pooled population is symmetrically balanced: $p=1/2$.

Secondly, one defines the probability laws $F_k$'s by means of a procedure similar to that proposed in \cite{ASS00}, where the locations and the magnitude of the coefficients are  chosen according to a marked Poisson process. Hence, pairwise disjoint index sets $\E_1,\;\ldots,\; \E_K$ are randomly selected to characterize the $F_k$'s. Then, a random curve $X$ drawn from a given $F_k$ is defined as a sum of wavelets at scales $j\in \{j_0,\;\ldots,\; j_{\max}\}$ and positions $l\in\{0,\;\ldots,\; 2^j-1\}$ drawn in the index set $\E_k$, so that, we have with probability one,
\begin{equation}
X(t) = \sum_{(j,l)\in\E_k} \alpha_{j,l} \psi_{j,l}(t).
\end{equation}
Here, $\psi$ denotes a {\itshape{Beylkin}} wavelet, used to generate the toy example input signals. Observe also that, by construction, the supports $\X_1,\;\ldots,\; \X_K$ are pairwise disjoint, as the $\E_k$'s. \\

\subsection{Synthetic Data Simulation - Numerical Results}
\label{subsec:toyex}
Several experiments have been carried out mainly in order to shed some light on the advantage of local filtering, compared to global filtering, when implementing the {\sc TreeRank} approach. The experimental study also aims at analyzing the possible impact of the wavelet filtering parametrization on the performance of the {\sc Functional TreeRank} algorithm. The experimental design is summarized by the first $4$ columns of Table \ref{tab:res1}. In particular, we investigated the impact of the number of coefficients $N$ kept for prediction purpose; ranging from $N=102$ down to $N=10$, representing respectively $5\%$ and $0.5\%$ of the length of the input signals. Additionally, in order to explore the influence of the filtering method, different choices of wavelets and related parameters $j$ and $j_0$ involved in the filtering stage have been tested and compared: in experiments {\itshape{a1}} to {\itshape{a3}} (resp. {\itshape{b1}} to {\itshape{b3}}), the filtering parameters chosen are identical to those used for generating the input signals (marked with an asterisk), while in experiments {\itshape{a4}} to {\itshape{a8}} (resp. {\itshape{b4}} to {\itshape{b8}}), a different wavelet and/or wavelet scale parameters (\textit{i.e.} $j$ and $j_0$) are implemented. Eventually, different filtering schemes are proposed and compared over all the designed experiments: linear filtering \textit{vs.} non-linear filtering, either based on the $N$ wavelet coefficients with highest variance or else based on thresholded wavelet coefficients (\textit{cf} Remark \ref{rk:shrink}).

Hence, two algorithms are compared based on these synthetic datasets:
\begin{itemize}
\item the standard {\sc TreeRank} method described in \label{subsec:TreeRank}, taking as input the sampled signals preliminary filtered using wavelets once only, referred to as {\itshape{filtered {\sc TreeRank}}};
\item the {\sc Functional TreeRank} procedure, where an adaptive wavelet-based filtering is applied to the signals locally.
\end{itemize}

The partitioning rule {\sc{LeafRank}} involved in the implementation of both {\sc{TreeRank}} based algorithms is the {\sc{CART}} procedure, see \cite{cart84}. For all experiments, the number of terminal leaves of the subtrees has been limited to $8$, while that of the master ranking tree is less than $16$. 

These two approaches are compared based on two symmetrically balanced (with $p=1/2$) datasets, generated following the procedure previously explained: a test set containing $n_t=2000$ trajectories, used to evaluate the performance of the algorithms in terms of $\auc$, and a training set of size $n_l = 5000$. In order to compute averaged generalization performance, $B=50$ bootstrap samples, consisting of $n=2000$ trajectories, are drawn randomly from the training set following a standard bootstrap procedure (see \cite{BT94}). Resulting averaged $\auc$ and related standard deviation calculated on the test sample are summarized in Table \ref{tab:res1}. They are respectively denoted by $(\widehat{\auc}_{filt}, \widehat{\sigma}_{filt})$ when filtering is performed once, globally, before applying {\sc TreeRank} and by $(\widehat{\auc}_{func}, \widehat{\sigma}_{func})$ when {\sc Functional TreeRank} is implemented. \\

\begin{figure}[h!]
\begin{center}
\vspace{-0.5cm}
\begin{tabular}{cc}
\parbox{8cm}{
\begin{center}
\includegraphics[width = 8cm]{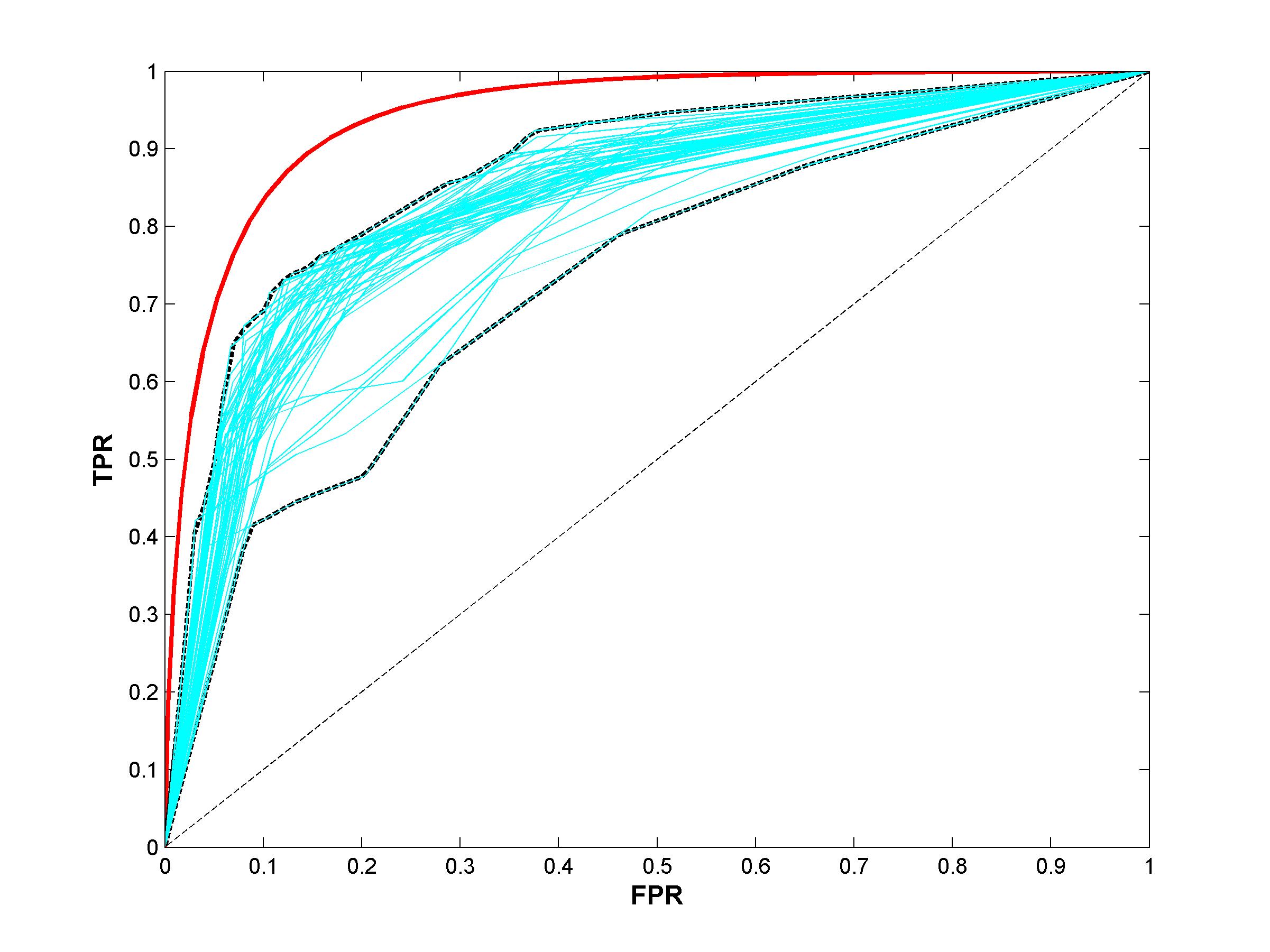}\\
{\small{a. {\itshape{Case a - Filtered {\sc TreeRank}}} }}\\
\end{center}}\\
\parbox{8cm}{
\begin{center}

\includegraphics[width = 8cm]{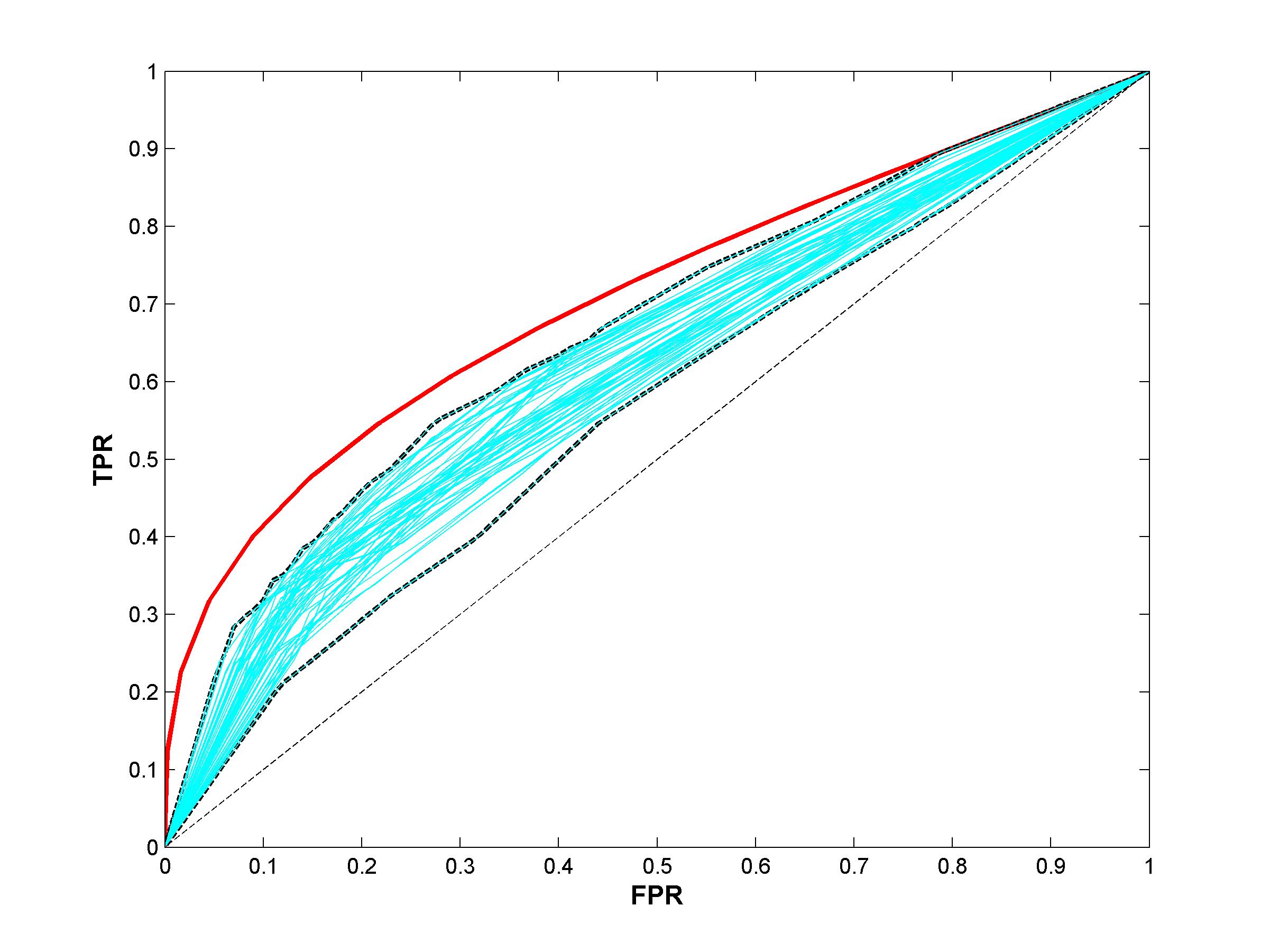}\\
{\small{b. {\itshape{Case b - Filtered {\sc TreeRank}}} }}\\
\end{center}}
\end{tabular}

\caption{{\small{Test $\roc$ envelopes for {\itshape{filtered}} {\sc TreeRank} over $100\%$ coefficients. Optimal $\roc^*$ are displayed in red, beams of test $\roc$ curves achieved by the {\sc TreeRank} based method are displayed in cyan dots and related enveloped in dotted black.}}\label{fig:rocs100}}
\end{center}
\end{figure}

As a first go, Fig. \ref{fig:rocs100} displays the results achieved by the {\itshape{filtered}} version of the {\sc TreeRank} algorithm, taking as input the wavelet-based filtered signals  without {\itshape{a priori}} selection of the coefficients; {\itshape{i.e.}} displayed test $\roc$ curves assess the performance of the algorithm taking into account $100\%$ ($N=2048$) of the wavelet coefficients. As can observe in Fig. \ref{fig:rocs100}, {\itshape{filtered}} {\sc TreeRank} achieves quite good results in terms of $\auc$ when taking into account $100\%$ of the computed coefficients. Indeed, the averaged test $\auc$ achieved raises up to $0.83$ with standard deviation of $0.02$ for {\itshape{Case a}}, and to $0.66$ with standard deviation of $0.01$ for {\itshape{Case b}}.

Based on these figures, we now analyze the numerical results for experiments {\itshape{a1}} to {\itshape{a8}} and {\itshape{b1}} to {\itshape{b8}}, summarized in Table \ref{tab:res1}. They highlight several points. First, these results emphasize the advantages of the {\sc Functional TreeRank} approach. Indeed, we observe that adaptive local filtering generally permits to improve significantly the performance of the {\sc{TreeRank}} algorithm, compared to the globally filtered approach. In particular, we can see that {\sc Functional TreeRank} achieves the best results when the wavelet coefficients are adaptively selected in a nonlinear manner. More specifically, local filtering combined with a nonlinear selection of wavelet coefficients with highest variance yields up to $10\%$ averaged increase in terms of $\auc$ criterion.

In addition, figures speak volume when the number of selected wavelet coefficients decreases significantly; yet, up to a certain point. Indeed, the performance of the {\sc TreeRank} based approaches remain comparable while the percentage of wavelet coefficients used for learning is above $5\%$. For most experiments, the $\auc$ values summarized in Table \ref{tab:res1} remain very close to those achieved by {\itshape filtered} {\sc TreeRank} and {\sc Logistic Regression} with $100\%$ of wavelet coefficients. Significant differences appear when this percentage decreases under $5\%$, see for example experiments $a2$, $a3$ and $b2$. However, when the number of coefficients kept for learning is much too low to represent properly input data, the advantages of {\sc Functional TreeRank} becomes naturally less obvious (e.g. experiment $b3$).

\begin{figure}[h!]
\begin{center} 
\includegraphics[width = 8cm]{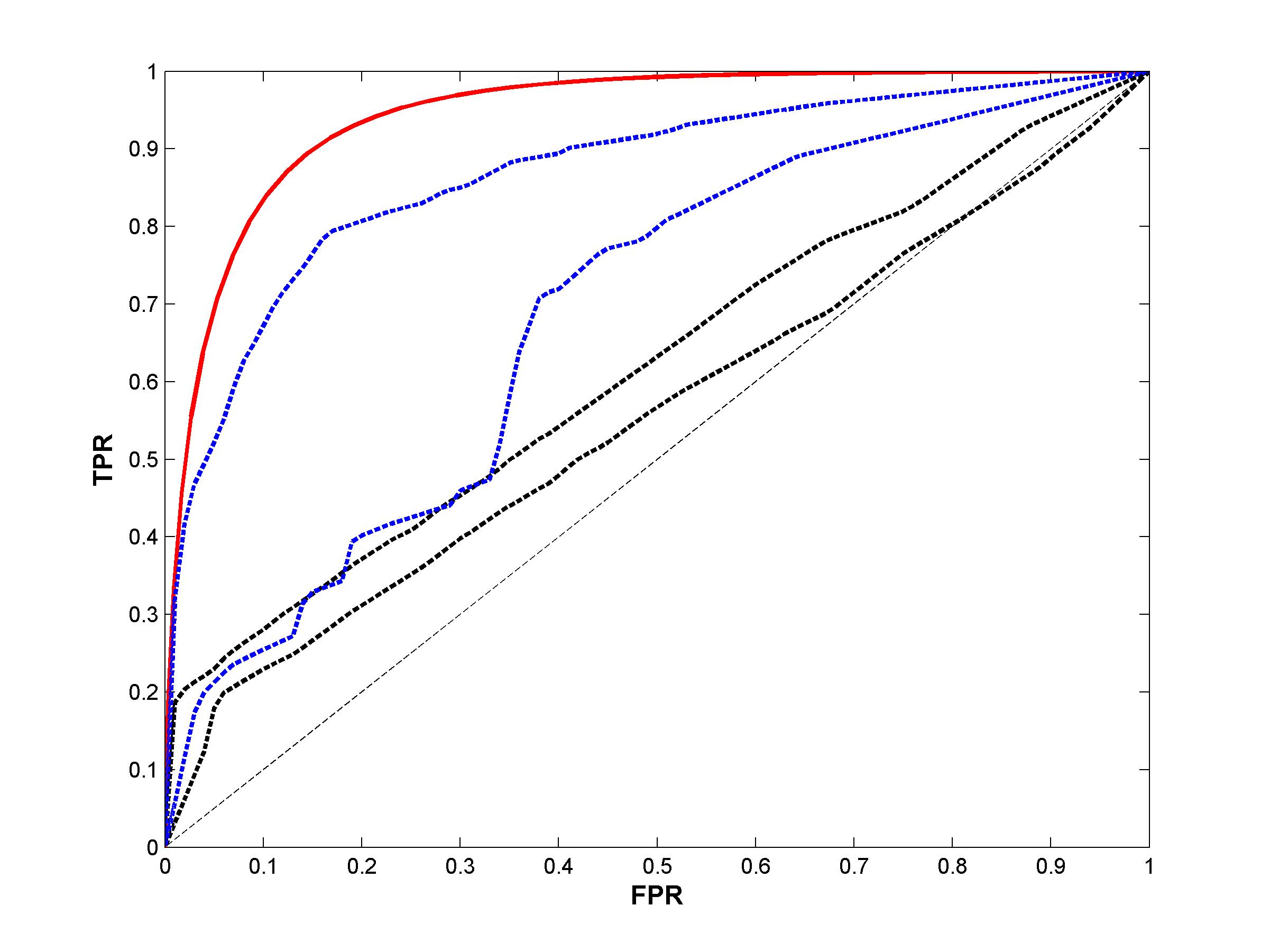}\\
\caption{{\itshape{Case a - Exp a3}} - $\roc$ Envelopes. {\small{Optimal $\roc^*$ in red, test $\roc$ envelopes achieved by {\itshape filtered {\sc TreeRank}} combined with a preliminary selection of $5\%$ wavelet coefficients and by {\sc Functional TreeRank} in dotted black and dotted blue respectively.}}\label{fig:corsa}}
\end{center}
\end{figure}

Additionally, when the number of coefficients decreases, the filtering scheme appears to have a profound effect on the performance of the {\itshape filtering-based} learning methods compared. Depending on the filtering procedure, we observe that both the impact of adaptive local filtering and the performance evaluated in terms of $\auc$ are significantly modified. In particular, the performance achieved by algorithms based on linear filtering significantly decreases when the percentage of kept coefficients gets lower than $5\%$. The same observation can be made for nonlinear filtering via thresholding, to a lesser degree however. The $\roc$ envelopes displayed in Fig. \ref{fig:corsa} clearly shows the difference between global {\itshape a priori} filtering and adaptive local filtering in terms of ranking performance on experiment $a3$. In particular, even if {\sc Functional TreeRank} appears to be slightly less stable (larger $\roc$ envelope) in this experiment, it is still much more accurate in terms of $\auc$ than the version based on a preliminary filtering of the training data.\\

In addition, experiments $a4$ to $a8$ and $b4$ to $b8$ also give some information on the impact of the wavelet-based filtering parameters. In particular, it clearly appears that the choice of the scale parameter $j$ may have a strong impact on the performance of the algorithm. In experiments $a7$ and $b7$, all the considered methods present significantly decreased performance due to a sub-optimal value of parameter $j$. Nevertheless, experiments $a8$ and $b8$ tend to show that the influence of this parameter remains lower than that of the wavelet chosen. Indeed, with the same $j$ values but a different wavelet to represent input signals, the performances reached become comparable again to those attained with the optimal parameters (see experiments $a1$ and $b1$). However, the displayed experiments do not permit to arrive at a general conclusion regarding the impact of the wavelet used for filtering the data. On the one hand, performances achieved in experiments $a4$ and $a5$ (for highest variance based filtering), based on a Daubechies wavelet decomposition, appear to be greatly improved in comparison with experiments $a1$. On the other hand, no significant difference appears in the experiments related to {\itshape Case b}. 

\newpage

\begin{figure}[t!]
\hspace{-0.5cm}
\vspace{0.5cm}
\begin{sideways}


\centering


\begin{tabular}{|c|c|c|c|c|c|c|c|c|c|c|}

\hline

 {}&\multicolumn{4}{|c|}{\multirow{2}{*}{Filtering}} & \multicolumn{2}{|c|}{\multirow{2}{*}{Lin. Coef. Sel.}} & \multicolumn{2}{|c|}{\multirow{2}{*}{ Non-lin. coef. Sel.}} & \multicolumn{2}{|c|}{\multirow{2}{*}{ Coef. Sel. via Thresholding}} \\ 

 &\multicolumn{4}{|c|}{} & \multicolumn{2}{|c|}{ }&\multicolumn{2}{|c|}{ } &\multicolumn{2}{|c|}{ }  \\ \hline 

 Experiment &\multirow{2}{*}{Wavelet} & \multirow{2}{*}{$j$} &\multirow{2}{*}{$j0$} & $N$  &$\widehat{\auc}_{filt}$&$\widehat{\auc}_{func}$ &$\widehat{\auc}_{filt}$&$\widehat{\auc}_{func}$ &$\widehat{\auc}_{filt}$&$\widehat{\auc}_{func}$ \\& & & & {\small{($\%$)}} & {\small{($\widehat{\sigma}_{filt}$)}} &	{\small{($\widehat{\sigma}_{func}$)}} & {\small{($\widehat{\sigma}_{filt}$)}} &	{\small{($\widehat{\sigma}_{func}$)}}& {\small{($\widehat{\sigma}_{filt}$)}} &	{\small{($\widehat{\sigma}_{func}$)}}\\ \hline \hline

{\itshape{Exp. a1}}	& Beylkin {\small{(*)}}	&	$11$ {\small{(*)}}	&	$1$ {\small{(*)}}	&	$102$ {\small{($5\%$)}} &	$0.82$ {\small{($\pm 0.01$)}}	&	$0.82$ {\small{($\pm 0.01$)}}	&	$0.78$ {\small{($\pm 0.02$)}}	&	${\textbf{0.83}}$ {\small{($\pm {\textbf{0.03}}$)}} &	$0.72$ {\small{($\pm 0.01$)}} &	$0.77$ {\small{($\pm 0.03$)}}\\ \hline

{\itshape{Exp. a2}}		& Beylkin {\small{(*)}}	&	$11$ {\small{(*)}}	&	$1$ {\small{(*)}}	&	$20$ {\small{($1\%$)}} &		$0.61$ {\small{($\pm 0.01$)}}	&	$0.61$ {\small{($\pm 0.01$)}}	&	$0.59$ {\small{($\pm 0.01$)}}	&	${\textbf{0.78}}$ {\small{($\pm {\textbf{0.04}}$)}} &	$0.72$ {\small{($\pm 0.01$)}} &	${\textbf{0.77}}$ {\small{($\pm {\textbf{0.03}}$)}}\\ \hline 

{\itshape{Exp. a3}}		& Beylkin {\small{(*)}}	&	$11$ {\small{(*)}}	&	$1$ {\small{(*)}}	&	$10$ {\small{($0.5\%$)}} &		$0.52$ {\small{($\pm 0.01$)}}	&	$0.52$ {\small{($\pm 0.01$)}}	&	$0.59$ {\small{($\pm 0.01$)}}	&	$0.74$ {\small{($\pm 0.05$)}} &	${\textbf{0.73}}$ {\small{($\pm {\textbf{0.01}}$)}} &	$0.73$ {\small{($\pm 0.04$)}}\\ \hline \hline

{\itshape{Exp. a4}}	&	Daubechies $12$	&	$11$ {\small{(*)}}	&	$1$ {\small{(*)}}	&	$102$ {\small{($5\%$)}}&	$0.85$ {\small{($\pm 0.01$)}}	&	$0.85$ {\small{($\pm 0.01$)}}	&	$0.89$ {\small{($\pm 0.01$)}}	&	$0.89$ {\small{($\pm 0.01$)}} &	$0.88$ {\small{($\pm 0.01$)}} &	$0.88$ {\small{($\pm 0.01$)}}\\ \hline

{\itshape{Exp. a5}}		&	Daubechies $4$	&	$11$ {\small{(*)}}	&	$1$ {\small{(*)}}	&	$102$ {\small{($5\%$)}}&	${\textbf{0.87}}$ {\small{($\pm {\textbf{0.01}}$)}}	&	${\textbf{0.87}}$ {\small{($\pm {\textbf{0.01}}$)}}	&	${\textbf{0.89}}$ {\small{($\pm {\textbf{0.01}}$)}}	&	${\textbf{0.89}}$ {\small{($\pm {\textbf{0.01}}$)}} &	$0.71$ {\small{($\pm 0.01$)}} &	$0.77$ {\small{($\pm 0.03$)}}\\ \hline

{\itshape{Exp. a6}}		&	Beylkin {\small{(*)}}	&	$11$ {\small{(*)}}	&	$4$ &	$102$ {\small{($5\%$)}}&	$0.82$ {\small{($\pm 0.02$)}}	&	$0.82$ {\small{($\pm 0.02$)}}	&	$0.78$ {\small{($\pm 0.02$)}}	&	${\textbf{0.89}}$ {\small{($\pm {\textbf{0.01}}$)}} &	$0.67$ {\small{($\pm 0.01$)}} &	$0.69$ {\small{($\pm 0.06$)}}\\ \hline

{\itshape{Exp. a7}}		&	Beylkin {\small{(*)}}	&	$8$ 	&	$1$ {\small{(*)}} &	$102$ {\small{($5\%$)}}&	$0.52$ {\small{($\pm 0.02$)}}	&	$0.52$ {\small{($\pm 0.02$)}}	&	$0.72$ {\small{($\pm 0.01$)}}	&	${\textbf{0.83}}$ {\small{($\pm {\textbf{0.03}}$)}} &	$0.51$ {\small{($\pm 0.01$)}} &	$0.55$ {\small{($\pm 0.04$)}}\\ \hline

{\itshape{Exp. a8}}		&	Daubechies $12$	&	$8$ 	&	$4$ &	$102$ {\small{($5\%$)}}&	$0.82$ {\small{($\pm 0.01$)}}	&	$0.82$ {\small{($\pm 0.01$)}}	&	${\textbf{0.85}}$ {\small{($\pm {\textbf{0.01}}$)}} &	${\textbf{0.85}}$ {\small{($\pm {\textbf{0.01}}$)}} &	$0.78$ {\small{($\pm 0.01$)}} & $0.78$ {\small{($\pm 0.01$)}} \\ \hline \hline

{\itshape{Exp. b1}}		& Beylkin {\small{(*)}}	&	$11$ {\small{(*)}}	&	$1$ {\small{(*)}}	&	$102$ {\small{($5\%$)}} &	$0.64$ {\small{($\pm 0.02$)}}	&	$0.64$ {\small{($\pm 0.02$)}}	&	$0.63$ {\small{($\pm 0.02$)}}	&	$0.64$ {\small{($\pm 0.02$)}} &	$0.63$ {\small{($\pm 0.02$)}} &	$0.64$ {\small{($\pm 0.02$)}}\\ \hline

{\itshape{Exp. b2}}		& Beylkin {\small{(*)}}	&	$11$ {\small{(*)}}	&	$1$ {\small{(*)}}	&	$20$ {\small{($1\%$)}} &	${\textbf{0.56}}$ {\small{($\pm {\textbf{0.01}}$)}}	&	${\textbf{0.56}}$ {\small{($\pm {\textbf{0.01}}$)}}	&	$0.50$ {\small{($\pm 0.01$)}}	&	$0.58$ {\small{($\pm 0.05$)}} &	${\textbf{0.57}}$ {\small{($\pm {\textbf{0.01}}$)}} &	${\textbf{0.60}}$ {\small{($\pm {\textbf{0.03}}$)}}\\ \hline 

{\itshape{Exp. b3}}		& Beylkin {\small{(*)}}	&	$11$ {\small{(*)}}	&	$1$ {\small{(*)}}	&	$10$ {\small{($0.5\%$)}} &	$0.56$ {\small{($\pm 0.01$)}}	&	$0.56$ {\small{($\pm 0.01$)}}	&	$0.51$ {\small{($\pm 0.01$)}}	&	$0.55$ {\small{($\pm 0.04$)}} &	$0.56$ {\small{($\pm 0.01$)}} &	$0.57$ {\small{($\pm 0.03$)}}\\ \hline \hline

{\itshape{Exp. b4}}		&	Daubechies $12$	&	$11$ {\small{(*)}}	&	$1$ {\small{(*)}}	&	$102$ {\small{($5\%$)}}&	$0.68$ {\small{($\pm 0.01$)}}	&	$0.68$ {\small{($\pm 0.01$)}}	&	$0.67$ {\small{($\pm 0.01$)}}	&	$0.67$ {\small{($\pm 0.01$)}} &	$0.67$ {\small{($\pm 0.01$)}} &	$0.67$ {\small{($\pm 0.01$)}}\\ \hline

{\itshape{Exp. b5}}		&	Daubechies $4$	&	$11$ {\small{(*)}}	&	$1$ {\small{(*)}}	&	$102$ {\small{($5\%$)}}&	$0.67$ {\small{($\pm 0.01$)}}	&	$0.67$ {\small{($\pm 0.01$)}}	&	$0.66$ {\small{($\pm 0.01$)}}	&	$0.67$ {\small{($\pm 0.01$)}} &	$0.64$ {\small{($\pm 0.01$)}} &	$0.65$ {\small{($\pm 0.02$)}}\\ \hline

{\itshape{Exp. b6}}		&	Beylkin {\small{(*)}}	&	$11$ {\small{(*)}}	&	$4$ &	$102$ {\small{($5\%$)}}&	${\textbf{0.64}}$ {\small{($\pm {\textbf{0.02}}$)}}	&	${\textbf{0.64}}$ {\small{($\pm {\textbf{0.02}}$)}}	&	${\textbf{0.63}}$ {\small{($\pm {\textbf{0.02}}$)}}	&	${\textbf{0.64}}$ {\small{($\pm {\textbf{0.02}}$)}} &	$0.50$ {\small{($\pm 0.01$)}} &	$0.52$ {\small{($\pm 0.02$)}}\\ \hline

{\itshape{Exp. b7}}		&	Beylkin {\small{(*)}}	&	$8$ 	&	$1$ {\small{(*)}} &	$102$ {\small{($5\%$)}}&	${\textbf{0.56}}$ {\small{($\pm {\textbf{0.01}}$)}}	&	${\textbf{0.56}}$ {\small{($\pm {\textbf{0.01}}$)}}	&	$0.50$ {\small{($\pm 0.01$)}}	&	$0.58$ {\small{($\pm 0.05$)}} &	$0.51$ {\small{($\pm 0.01$)}} &	$0.51$ {\small{($\pm 0.01$)}}\\ \hline

{\itshape{Exp. b8}}		&	Daubechies $12$	&	$8$ 	&	$4$ &	$102$ {\small{($5\%$)}}&	$0.66$ {\small{($\pm 0.01$)}}	&	$0.66$ {\small{($\pm 0.01$)}}	&	$0.66$ {\small{($\pm 0.01$)}}	&	$0.66$ {\small{($\pm 0.01$)}} &	$0.66$ {\small{($\pm 0.01$)}} &	$0.66$ {\small{($\pm 0.01$)}}\\ \hline
\end{tabular}

\end{sideways}
\caption{{\itshape{Case a and b}} - global filtering {\itshape{versus}} local filtering.\label{tab:res1}}
\end{figure}

\subsection{Real dataset - Experiment Design and Numerical Results}


The experiments presented in this subsection are based on a real dataset, used as a benchmark in the Brain Computer Interface (BCI in abbreviated form) community. Typical BCI datasets consist of a number of brain signals collected by several sensors scattered over the head of one or several subjects. These signals reflect the brain activity when exposed to an external stimulus, \textit{e.g.} audio, visual or intellectual stimulation. One of the main objectives in BCI design is to perform brain signal analysis so as to identify and extract relevant features characterizing a given stimulus. \\

The dataset considered in this experiment gathers $n=216$ observations; each of these consists of a collection of time series measured by $nc=203$ sensors scattered over the head of given subjects. Each time series is composed of $n_t=121$ time points. Four different stimuli have been tested on the subjects: two audio stimulations and two visual stimulations namely. For each of these, $54$ samples are observed. The analysis of the available dataset thus corresponds to a multi-class problem. Yet, data samples can be grouped, audio stimulation on the one hand, visual ones on the other hand, so as to bring us back to the bipartite setting. Two different configurations are considered:
\begin{itemize}
\item {\itshape Case c}: a dataset of $n=108$ observations is analyzed, with $54$ samples subject to a given audio stimulus and the other $54$ to a given visual stimulus\\
\item {\itshape Case d}: the full dataset is analyzed but both audio (resp. visual) stimulations are considered jointly, so as to bring us back to the bipartite setting.
\end{itemize}
As for previous experiments, different wavelet decompositions are tried and tested, as detailed in the table of Fig. \ref{tab:configbench}. For each experiment, $2^j$ wavelet coefficients are computed per brain signal; thus, each observation, originally stored as a matrix of size $121\times 203$, is reresented by a vector of length ranging from $1624$ to $3248$, depending on the value chosen for parameter $j$.

\begin{figure}[!ht]
\centering

\begin{tabular}{|c|c|c|c|}

\hline

\multirow{3}{*}{Experiments}  &\multicolumn{3}{|c|}{\multirow{2}{*}{Filtering Parameters}} \\ 

&\multicolumn{3}{|c|}{} \\ 
\cline{2-4}
& Wavelet &$j$ &$j_0$ \\ 

\hline \hline

{\itshape{Exp. 1}}	& Daubechies $20$	&	$3$	&	$0$	\\ \hline

{\itshape{Exp. 2}}		&Daubechies $20$	&	$3$	&	$2$	\\ \hline 

{\itshape{Exp. 3}}		& Daubechies $20$	&	$4$	&	$0$	  \\ \hline 

{\itshape{Exp. 4}}	&	Daubechies $20$	&	$4$	&	$2$	\\ \hline \hline

{\itshape{Exp. 5}}		&	Coiflet $2$	&	$3$	&	$0$	\\ \hline

{\itshape{Exp. 6}}		&	Coiflet $2$	&	$3$	&	$2$	\\ \hline 

{\itshape{Exp. 7}}		&	Coiflet $2$		&	$4$	&	$0$	  \\ \hline 

{\itshape{Exp. 8}}		&	Coiflet $2$		&	$4$	&	$2$	\\ \hline \hline

{\itshape{Exp. 9}}		&	Symmlet $10$	&	$3$	&	$0$	\\ \hline

{\itshape{Exp. 10}}		&	Symmlet $10$	&	$3$	&	$2$	\\ \hline 

{\itshape{Exp. 11}}		&	Symmlet $10$	&	$4$	&	$0$	  \\ \hline 

{\itshape{Exp. 12}}		&	Symmlet $10$		&	$4$	&	$2$	\\ \hline 
\end{tabular}\\
\vspace{0.25cm}
\caption{ {\small{ Experiments design - Filtering parametrization }}\label{tab:configbench}}
\end{figure}
\newpage
In order to discriminate and characterize both types of stimulations, the {\sc Functional TreeRank} algorithm is performed, with the same {\sc LeafRank} partitioning rule used as for the synthetic toy example, based on the {\sc CART} procedure. As previously, the maximum number of terminal leaves of the subtrees has been settled to $8$, while the master ranking tree size is limited to $16$ terminal leaves. The wavelet coefficients are selected by {\sc Functional TreeRank} through non-linear filtering of the input data, based on the $5\%$ wavelet coefficients with highest variance (see \ref{subsec:nonlin}). \\

Different validation procedure are considered in order to evaluate the generalization capacity of the ouput tree-structured scoring rule. 

Firstly, a $V$-fold cross-validation is performed. 
In {\itshape Case c}, where the dataset is quite small ($108$ samples), a $4$-fold cross-validation is performed, while $6$-folds are considered in {\itshape Case d}. Averaged $\auc$ and related standard deviations are summarized in the table of Fig. \ref{tab:bci}, in the column {\itshape V-fold}. 

Secondly, a classical bootstrap-based validation process is considered, similar to that we described in \ref{subsec:toyex}. Therefore, for each experiment, the original datasets are firstly divided into a learning sample, containing $2/3$ of the dataset, and the complementary test sample. Then, $B=30$ bootstrap samples, of size $nl=200$, are drawn with replacement from the original learning sample. Based on the latter, $30$ scoring rules are build by the {\sc Functional TreeRank} algorithm and then evaluated on the test set. Resulting averaged $\auc$ and related standard deviations are summarized in Fig. \ref{tab:bci}, in the column {\itshape Boot}.



\begin{figure*}



\begin{center}

\begin{tabular}{|c|c|c|c|c|}

\hline

 {}& \multicolumn{2}{|c|}{\multirow{2}{*}{\itshape Case c}} & \multicolumn{2}{|c|}{\multirow{2}{*}{\itshape Case d}}  \\ 

 &\multicolumn{2}{|c|}{ } &\multicolumn{2}{|c|}{ }\\ \hline 

 Experiment & $\widehat{\auc}_{V-fold}$&$\widehat{\auc}_{Boot}$ & $\widehat{\auc}_{V-fold}$&$\widehat{\auc}_{Boot}$  \\
 &	{\small{($\widehat{\sigma}_{V-fold}$)}} & {\small{($\widehat{\sigma}_{Boot}$)}} 
 &	{\small{($\widehat{\sigma}_{V-fold}$)}} & {\small{($\widehat{\sigma}_{Boot}$)}} \\ \hline \hline

{\itshape{Exp. 1}}	& $0.68$ {\small{($\pm 0.05$)}}	&	$0.61$ {\small{($\pm 0.06$)}}	& $0.75$ {\small{($\pm 0.09$)}}	&	$0.75$ {\small{($\pm 0.07$)}}\\ \hline

{\itshape{Exp. 2}}	& $0.74$ {\small{($\pm 0.09$)}}	&	$0.65$ {\small{($\pm 0.08$)}}	& $0.86$ {\small{($\pm 0.12$)}}	&	$0.84$ {\small{($\pm 0.05$)}}\\ \hline

{\itshape{Exp. 3}}	& $0.80$ {\small{($\pm 0.14$)}}	&	$0.72$ {\small{($\pm 0.1$)}}	& $0.87$ {\small{($\pm 0.08$)}}	&	$0.86$ {\small{($\pm 0.04$)}}\\ \hline

{\itshape{Exp. 4}}	& $0.72$ {\small{($\pm 0.06$)}}	&	$0.71$ {\small{($\pm 0.1$)}}	& $0.87$ {\small{($\pm 0.08$)}}	&	$0.84$ {\small{($\pm 0.05$)}}\\ \hline\hline

{\itshape{Exp. 5}}		& $0.66$ {\small{($\pm 0.05$)}}	&	$0.63$ {\small{($\pm 0.07$)}}	& $0.74$ {\small{($\pm 0.06$)}}	&	$0.90$ {\small{($\pm 0.04$)}}\\ \hline

{\itshape{Exp. 6}}		& $0.61$ {\small{($\pm 0.07$)}}	&	$0.63$ {\small{($\pm 0.08$)}}	& $0.82$ {\small{($\pm 0.08$)}}	&	$0.88$ {\small{($\pm 0.06$)}}\\ \hline

{\itshape{Exp. 7}}		& $0.89$ {\small{($\pm 0.07$)}}	&	$0.72$ {\small{($\pm 0.06$)}}	&$0.96$ {\small{($\pm 0.03$)}}	&	$0.88$ {\small{($\pm 0.04$)}}\\ \hline

{\itshape{Exp. 8}}		& $0.85$ {\small{($\pm 0.04$)}}	&	$0.69$ {\small{($\pm 0.06$)}}& $0.96$ {\small{($\pm 0.06$)}}	&	$0.87$ {\small{($\pm 0.06$)}}	\\ \hline \hline

{\itshape{Exp. 9}}		& $0.65$ {\small{($\pm 0.05$)}}	&	$0.61$ {\small{($\pm 0.06$)}}& $0.74$ {\small{($\pm 0.09$)}}	&	$0.75$ {\small{($\pm 0.07$)}}	\\ \hline

{\itshape{Exp. 10}}		& $0.74$ {\small{($\pm 0.09$)}}	&	$0.66$ {\small{($\pm 0.07$)}}	&$0.86$ {\small{($\pm 0.12$)}}	&	$0.82$ {\small{($\pm 0.06$)}}\\ \hline

{\itshape{Exp. 11}}		& $0.80$ {\small{($\pm 0.14$)}}	&	$0.84$ {\small{($\pm 0.07$)}}& $0.87$ {\small{($\pm 0.08$)}}	&	$0.87$ {\small{($\pm 0.04$)}}	\\ \hline

{\itshape{Exp. 12}}		& $0.77$ {\small{($\pm 0.12$)}}	&	$0.79$ {\small{($\pm 0.11$)}}	& $0.87$ {\small{($\pm 0.09$)}}&	$0.87$ {\small{($\pm 0.04$)}}\\ \hline 
\end{tabular}
\vspace{0.25cm}
\caption{ {\small{ {\itshape Case c and d} - {\sc Functional TreeRank} algorithm performances}}\label{tab:bci}}
\end{center}
\end{figure*}

We observe that {\sc Functional TreeRank} leads to quite good results on this benchmark dataset. This should be however tempered, considering the relativly high standard deviation values, ranging approximately from $5\%$ to $18\%$ of the $\auc$ value, due to the small number of samples available. Nevertheless, the performance tends to be better on {\itshape Case d}, both in terms of $\auc$ and standard deviation (at least for the bootstrap validation process). Besides, it also appears that choice of the filtering parametrization highly impacts the algorithms' performances. In particular, on {\itshape Case c}, the choice of of higher value ($4$) for the $j$ parameter enables to improve the ranking accuracy.

\subsection{Discussion}
The experimental results obtained deserve some comments. In order to gain insight into the reason why {\sc Functional TreeRank} outperforms its competitors here, it is important to keep in mind that, in Functional Classification, the goal is to recover a single specific level set of the regression function $\eta$ (the set of instances $x$ s.t. $\eta(x)>1/2$). A good filtering will be thus evaluated according to its capacity to describe this particular set, to capture the regularity/geometry of its frontier. By contrast, ranking aims at recovering the (monotonous) collection of all regression level sets (without necessarily knowing the corresponding levels). The adequate way of filtering the data for this purpose may (possibly heavily) depend on the level considered. By nature, because it boils down to recover an adaptively selected sub-collection of (bi)level regression sets in a recursive fashion, the {\sc TreeRank} approach permits to implement filtering \textit{locally}, depending on the bilevel set to recover. As regards to {\sc TreeRank}'s competitors, RankBoost \cite{FISS03} or RankSVM \cite{HGBSO98} for instance, we highlight the fact that it is far from easy to extend this principle to these alternative methods, mostly based on pairwise classification. Straightforward applications of these approaches in the functional situation would naturally consist in implementing the latter based on signals filtered once and only once. 
As illustrated by the numerical examples given above, bilevel sets corresponding to high levels can be well approximated through representations based on high frequency components for instance, while those corresponding to low levels can be captured by means of low frequency terms. The ability of wavelets to adapt to unknown degree of smoothness is thus also a key to ranking performance in the functional context. In brief, {\sc Functional TreeRank} is more efficient not only because it uses more features, but primarily and especially because it involves the right features for the right subproblem (recovery of a specific bilevel set). Incidentally, we point out that augmenting significantly the number of features (in a preliminary filtering stage) would dramatically increase the variance term and, consequently, would deteriorate the performance, as underlined by the short discussion about learning rates in the end of Section 3.

\section{Conclusion}
In this article, bipartite ranking is tackled from the angle of Functional Data Analysis for the first time. A filtering approach based on wavelet analysis is promoted and the decrease of optimal $\auc$ caused by the filtering stage is related to the distortion rate achieved by the wavelet approximants considered. A novel learning algorithm, based on ordered recursive partitioning of the path space, is proposed in this context, where adaptive filtering is implemented \textit{locally}, at each splitting step, in order to handle situations where the form of the best $N$-term approximant for the input signal $X$ highly depends on the values taken by the regression function $\eta(X)$.  Convincing experimental results are reported in illustrative examples, offering promising perspectives for the use of the approach promoted here in specific applications, such as anomaly detection, medical diagnosis support or credit-scoring, which shall undoubtedly rely more and more commonly on functional datasets in the Big Data era.

\appendices
\section{Proof of Theorem \ref{thm:approx2}}
Using the orthonormality of the wavelet basis, observe that the expected distortion, measured in the $\mathcal{L}_2$ sense, can be written as
$$
\mathbb{E}[\vert\vert X-\widetilde{\mathcal{P}}_N X\vert\vert^2]=\sum_{(j,k)\notin \mathcal{I}_N}\mathbb{E}[\beta_{j,k}^2].
$$
Notice in addition that condition (\ref{eq:constr}) implies that the rearrangement $(\beta_{j(m),k(m)})_{m\geq 1}$ of the wavelet coefficients in decreasing order of their expected square fulfills the decay property:
$$
\mathbb{E}[\beta_{j(m),k(m)}^2]^{p/2}\leq c_0 \cdot 1/m,
$$
for some constant $c_0<+\infty$. One may then write
\begin{eqnarray*}
\sum_{(j,k)\notin \mathcal{I}_N}\mathbb{E}[\beta_{j,k}^2]=\sum_{m>N}\mathbb{E}[\beta_{j(m),k(m)}^2]\leq c\cdot N^{1-2/p},
\end{eqnarray*}
for some finite constant $c$.

\section{Proof of Proposition \ref{prop:bias}  (Sketch of)}
Consider first assertion $(i)$. Following in the footsteps of the argument of Lemma 32.3 in \cite{DGL96}, observe that: $\forall j\geq j_0$,
\begin{eqnarray*}
\auc^*-\auc^*_j&\leq& \auc(\eta)-\auc(\eta_j)\\
&\leq & \frac{1}{p(1-p)}\mathbb{E}\left[ \vert \eta(X)-\eta(\mathcal{P}_jX) \vert\right],
\end{eqnarray*}
where $\eta_j(X)\overset{def}{=}\eta(\mathcal{P}_jX )$ and the second bound results from Proposition $8$ in \cite{CV09ieee}.
Now the desired result follows from Theorem \ref{thm:approx1} combined with the Lipschitz assumption. 

\par Assertion $(ii)$ can be proved in a similar manner, appealing to Theorem \ref{thm:approx2}.

\section{Proof of Proposition \ref{prop:select} (Sketch of)}
The proof relies on a version of the Vapnik-Chervonenkis inequality for the $\auc$ in the $N$-dimensional setting used in \cite{CDV09} (see Proposition 6's argument therein, see also \cite{CLV08}), combined with the fact the upper bound it provides for the (expected) uniform deviation between empirical and theoretical $\auc$'s is distribution free and thus leads here to a bound that does not depend on the filter $\pi_{\mathcal{I}_N}$ (namely, on $\pi_{\mathcal{I}_N}(X)$'s distribution) but on $\S_N$'s complexity solely. Precisely, it yields
\begin{multline*}
\mathbb{E}\left[\sup_{S\in \S_N}\vert \widehat{\auc}(S\circ \pi_N)-\auc(S\circ \pi_N) \vert\right]\leq\\
 4\sqrt{\frac{V_N\log(n+1)+\log 2}{n}}.
\end{multline*}
The oracle inequality then straightforwardly follows from the argument of Theorem 1.20 in \cite{Lug02}.

\section*{Acknowledgment}
The authors gratefully thank Prof. N. Vayatis for helpful and encouraging comments and Prof. A. Gramfort for providing them with the BCI dataset used in the present article.
\bibliography{References_ranking}
\bibliographystyle{IEEEtran}

\end{document}